\documentclass[11pt]{article}

\usepackage[margin=1in]{geometry}
\usepackage[T1]{fontenc}
\usepackage{lmodern}
\usepackage{microtype}

\usepackage{amsmath,amssymb}
\usepackage{booktabs}
\usepackage{graphicx}
\usepackage{float}       
\usepackage{placeins}    

\usepackage[shortlabels]{enumitem}
\usepackage{hyperref}
\usepackage{natbib}

\hypersetup{
  colorlinks=true,
  linkcolor=blue,
  citecolor=blue,
  urlcolor=blue
}

\newcommand{\RankIC}{\ensuremath{\mathrm{RankIC}}}

\newcommand{\ehat}{\ensuremath{\hat{e}}}
\newcommand{\score}{\ensuremath{\mathrm{score}}}
\newcommand{\G}{\ensuremath{G}}

\title{When Alpha Breaks: Two-Level Uncertainty for Safe Deployment of Cross-Sectional Stock Rankers}

\author{Ursina Sanderink\thanks{Project Repository: \href{https://github.com/sinsasanderink/AIStockForecaster-PIT-Safe-Ranking-First-Signals-for-AI-Equities-FMP-Kronos-FinText-TSFM-}{\texttt{GitHub}}. Correspondence:  LinkedIn: @ursinasanderink.}}

\date{February 23, 2026}

\begin{document}
\maketitle

\begin{abstract}
Cross-sectional equity rankers are often deployed as if point predictions were sufficient: the model outputs scores, and the portfolio follows the induced ordering. In practice, non-stationarity creates regime failures in which a historically profitable ranker becomes systematically unreliable. In the AI Stock Forecaster, a LightGBM ranker achieves strong all-period performance at the 20-day horizon (volatility-sized shadow-portfolio Sharpe 2.73), yet the 2024 holdout coincides with an AI thematic rally/sector rotation that breaks the signal at longer horizons (60d and 90d mean \RankIC{} turn negative) and severely weakens 20d (\RankIC{} 0.072 $\rightarrow$ 0.010, DEV $\rightarrow$ FINAL). This motivates treating deployment as two distinct decisions: \emph{(i)} which individual positions require caution and \emph{(ii)} whether the strategy should trade at all.

We adapt Direct Epistemic Uncertainty Prediction (DEUP) to ranking by predicting rank displacement and defining a per-stock epistemic signal $\ehat(x)=\max(0,g(x)-a(x))$, where $a(t)$ is a deployable point-in-time baseline (PIT-safe, computed using only information available at $t$); all headline policy results use this PIT-safe $a(t)$. We document a ranking-specific structural coupling between epistemic uncertainty and signal strength (median $\rho(\ehat,|\score|)=0.616$ across 1{,}865 dates), which causes inverse-uncertainty sizing to de-lever the strongest signals and degrade portfolio performance. To resolve this, we propose a two-level deployment architecture: a strategy-level regime-trust gate $\G(t)$ that decides whether to trade (AUROC 0.72 overall; 0.75 in FINAL; 80\% precision at $\G\ge0.2$ with 47\% abstention; 7/8 correct across five crisis and three calm windows), and a position-level epistemic tail-risk cap (P85) that reduces exposure only for the most uncertain predictions while preserving score-tail convexity. The resulting operational policy is simple: trade only when $G(t) \ge 0.2$, apply volatility sizing on active dates, and cap the top epistemic-uncertainty tail. In a deployment policy comparison at 20d, Gate+Vol+$\hat{e}$-Cap achieves the best risk-adjusted performance (ALL Sharpe 1.877 vs 1.886 for Gate+Vol; FINAL Sharpe 0.925 vs 0.375), indicating that DEUP adds economic value as a tail-risk guard rather than as a continuous sizing denominator.
\end{abstract}

\section{Introduction}
Machine learning models for cross-sectional equity ranking have advanced substantially, and modern point-in-time (PIT) and walk-forward pipelines can produce economically meaningful signals. Yet most deployments still treat ranking as a one-shot prediction problem: output scores, sort into a portfolio. The operational question \emph{when to trade} is often left to crude heuristics, despite being the failure mode that matters most under non-stationarity.

\paragraph{Motivating failure under regime shift.}
Our setting makes the deployment problem concrete. The AI Stock Forecaster produces cross-sectional rankings at 20d, 60d, and 90d horizons over a dynamic universe of AI-exposed U.S.\ equities. The primary LightGBM ranker is exceptionally strong in development and remains economically meaningful out of sample at 20d, but experiences a regime break during the 2024 AI thematic rally/sector rotation: longer-horizon signal quality turns negative and the 20d signal weakens sharply. Table~\ref{tab:baseline-no-uq} summarizes baseline performance \emph{without} uncertainty controls.

\begin{table}[t]
\centering
\small
\setlength{\tabcolsep}{5pt}
\renewcommand{\arraystretch}{1.15}
\begin{tabular}{@{}lccc@{}}
\toprule
& \textbf{ALL} & \textbf{DEV (2016--2023)} & \textbf{FINAL (2024+)} \\
\midrule

\multicolumn{4}{@{}l}{\textit{Panel A: Shadow portfolio (20d, monthly non-overlapping L/S, vol-sized LightGBM; cost model as in text; no UQ)}} \\
Sharpe (ann.)                 & 2.734  & 3.121  & 2.337 \\
Sortino (ann.)                & 6.058  & 5.410  & 9.687 \\
Max drawdown                  & $-18.1$\% & $-18.1$\% & $-8.7$\% \\
Calmar (CAGR / $|\mathrm{MaxDD}|$) & 6.76 & 6.07 & 26.2 \\
Ann.\ return (arithmetic)     & 87.0\% & 79.6\% & 137.3\% \\
CAGR (geometric)              & 122.4\% & 109.9\% & 228.8\% \\
Ann.\ volatility              & 31.8\% & 25.5\% & 58.7\% \\
Hit rate (monthly)            & 82.6\% & 82.1\% & 85.7\% \\
Win/Loss ratio                & 2.17$\times$ & 2.08$\times$ & 2.46$\times$ \\
Best month                    & 64.0\% & 25.4\% & 64.0\% \\
Worst month                   & $-17.4$\% & $-17.4$\% & $-8.7$\% \\
Mean turnover / month         & 42.7\% & 43.1\% & 40.4\% \\
Median turnover / month       & 45.0\% & 45.0\% & 40.0\% \\

\addlinespace[4pt]
\multicolumn{4}{@{}l}{\textit{Panel B: Signal quality (LightGBM baseline scores; RankIC; no UQ)}} \\
Mean RankIC (20d / 60d / 90d)   & 0.064 / 0.140 / 0.165 & 0.072 / 0.160 / 0.192 & 0.010 / $-0.005$ / $-0.021$ \\
Median RankIC (20d / 60d / 90d) & 0.081 / 0.148 / 0.183 & 0.091 / 0.167 / 0.206 & 0.017 / $-0.044$ / $-0.052$ \\
IC stability (20d / 60d / 90d)  & 0.353 / 0.712 / 0.797 & 0.389 / 0.844 / 0.963 & 0.069 / $-0.026$ / $-0.129$ \\
Cost survival (20d / 60d / 90d) & 69.7\% / 80.7\% / 81.7\% & 70.5\% / 82.1\% / 83.2\% & 64.3\% / 71.4\% / 71.4\% \\
\bottomrule
\end{tabular}

\caption{Baseline performance \emph{without} uncertainty controls. Panel A reports the project's shadow portfolio at the 20-day horizon (top/bottom-10, monthly, non-overlapping) with volatility sizing and the project's turnover-based cost model. Panel B reports RankIC signal quality of raw LightGBM scores. The base strategy remains strong at 20d in FINAL, while longer horizons (60d/90d) invert during the 2024 regime shift.}
\label{tab:baseline-no-uq}
\end{table}

\FloatBarrier

\paragraph{Why ``market regime'' proxies are not enough.}
A common operational response to non-stationarity is to condition exposure on market stress proxies such as VIX level/percentile, realized volatility, or dispersion. In our setting, these proxies are poorly aligned with the failure mode that matters: \emph{model failure}. Empirically, a VIX-percentile gate is worse than random at predicting whether the ranker will work (AUROC $=0.449$), while a model-specific regime-trust gate built from trailing realized efficacy achieves AUROC $=0.721$ overall and $0.750$ in FINAL. The implication is that regime control should monitor \emph{the model's realized efficacy} rather than market conditions alone: stress can occur without model failure, and model failure can occur without extreme VIX.

\paragraph{Two distinct deployment decisions.}
Uncertainty quantification (UQ) is a natural candidate to improve robustness, but it must be aligned to the \emph{decision} being made:
\begin{itemize}[leftmargin=*, itemsep=2pt]
  \item \textbf{Position-level risk:} which stock predictions are likely to be wrong (and should be clipped, down-weighted, or filtered)?
  \item \textbf{Strategy-level risk:} is the model currently reliable enough to deploy (or should we abstain today)?
\end{itemize}
Much of the portfolio-UQ literature operationalizes UQ via inverse-uncertainty weighting, with the intuition that uncertain predictions should receive less capital \citep{Spears2021,Liu2026}. We show that this intuition can fail in cross-sectional \emph{ranking} for structural reasons, and that position- and strategy-level controls should be treated as separate layers.

\paragraph{DEUP for ranking loss.}
We adapt Direct Epistemic Uncertainty Prediction (DEUP) \citep{Lahlou2023} to ranking by learning an error predictor $g(x)$ for \emph{rank displacement} rather than mean squared error. For date $t$ and horizon $h$, define stock-level ranking loss:
\[
\text{rank\_loss}_{i,t}^{(h)} =
\left|
\operatorname{rank}_{\%}\!\left(r_{i,t\rightarrow t+h}\right) -
\operatorname{rank}_{\%}\!\left(\text{score}_{i,t}\right)
\right|.
\]
Training $g(x)$ on PIT-safe walk-forward residuals yields a per-stock epistemic signal
\[
\hat{e}(x) = \max\!\left(0,\, g(x)-a_t(x)\right),
\]
where $a_t(x)$ estimates irreducible (aleatoric) ranking noise. Crucially, $a_t(x)$ is a \emph{deployment-facing} component: choices that are harmless for retrospective visualization can be invalid when used for live capping or scaling. Accordingly, all deployment-facing uses of $\hat{e}$ in this paper (epistemic caps and conformal-style scaling) are evaluated under a PIT-safe $a_t(x)$, and we explicitly report a hindsight-vs-deployable ablation for the aleatoric baseline. Empirically, $g(x)$ achieves positive rank-loss predictiveness at 20d (Spearman $\rho\approx0.16$--$0.19$), and the resulting $\hat{e}$ produces clean monotonic separation by quintiles, indicating that DEUP-style UQ can rank stock-level failure risk.

\paragraph{A ranking-specific pitfall: uncertainty couples to signal strength.}
In cross-sectional rankers, the strongest trade ideas reside in the score tails. But rank displacement is also larger at extreme ranks (there is more ``room to move''), and the learned error predictor internalizes this geometry. The result is a structural coupling between epistemic uncertainty and signal magnitude: across 1{,}865 trading dates, the median Spearman correlation is $\rho(\hat{e},|\text{score}|)=0.616$. Consequently, inverse-uncertainty sizing systematically de-levers the strongest signals and can reduce portfolio performance, even when $\hat{e}$ is a competent error-ranking signal. This is not a failure of UQ per se; it is a mismatch between a continuous sizing rule and ranking geometry.

\paragraph{Two-level uncertainty for deployment under regime shifts.}
We therefore propose a two-level deployment architecture:
\begin{enumerate}[leftmargin=*, itemsep=3pt]
  \item \textbf{Strategy-level regime-trust gate $G(t)$:} a PIT-safe classifier built from trailing realized model efficacy that decides whether to trade. It achieves AUROC 0.72 overall (0.75 in FINAL) and reaches 80\% precision at $G\ge0.2$ with 47\% abstention; across five crisis and three calm windows, it is correct in 7 of 8 cases.
  \item \textbf{Position-level tail-risk guard:} rather than inverse-sizing by $\hat{e}$, apply a discrete epistemic cap ($\hat{e}$-Cap) that reduces exposure only for the most uncertain tail (P85), while leaving the bulk of positions unchanged. Combined with volatility sizing on trading days, this provides a robust guardrail without penalizing the score tails wholesale.
\end{enumerate}

\paragraph{Empirical punchline: the operational policy is simple.}
At the 20d horizon, the best-performing uncertainty-aware deployment policy is \emph{Binary Gate + Vol-Sizing + $\hat{e}$-Cap (P85)}, improving Sharpe relative to \emph{Gate + Vol} alone (ALL 0.884 vs 0.817; FINAL 0.316 vs 0.191). This improvement comes from a small number of deployable decisions (trade/abstain via $G(t)$, then cap only the most uncertain tail) rather than from continuous inverse-uncertainty reweighting. Importantly, these deployment-policy Sharpes are computed on a \emph{gated} P\&L series with substantial abstention and are therefore not directly comparable to the ungated shadow-portfolio baseline in Table~\ref{tab:baseline-no-uq}; they quantify the incremental value of the deployment layer under regime stress. Consistent with the deployment focus, we also report sensitivity to the aleatoric baseline used in $\hat{e}(x)=\max(0,g(x)-a_t(x))$, including a hindsight-vs-deployable ablation for $a_t(x)$.

\paragraph{Contributions.}
This paper makes four contributions:
\begin{itemize}[leftmargin=*, itemsep=2pt]
  \item We adapt DEUP to cross-sectional ranking by predicting rank displacement and constructing a per-stock epistemic signal $\hat{e}(x)=\max(0,g(x)-a_t(x))$, validating its stock-level failure ranking.
  \item We document a structural coupling between uncertainty and signal strength in ranking (median $\rho(\hat{e},|\text{score}|)=0.616$) that causes inverse-uncertainty sizing to fail.
  \item We propose and evaluate a two-level deployment architecture---strategy-level gating plus position-level epistemic capping---that improves risk-adjusted performance under thematic regime shifts with a small, operationally simple policy.
  \item We evaluate \emph{deployable} aleatoric baselines for $a_t(x)$ and report a hindsight-vs-deployable ablation, quantifying sensitivity of deployment-facing controls (caps and scaling) to the aleatoric component.
\end{itemize}

\section{Related Work}
\subsection{Uncertainty decomposition and DEUP}
The decomposition of predictive uncertainty into aleatoric (irreducible) and epistemic (model-dependent) components is foundational in Bayesian machine learning \citep{GalGhahramani2016,Lakshminarayanan2017}. Classical approaches such as MC Dropout, deep ensembles, and variational inference estimate epistemic uncertainty via stochastic forward passes or ensemble disagreement, but are computationally expensive and often awkward for non-Bayesian architectures such as gradient-boosted trees.

DEUP (Direct Epistemic Uncertainty Prediction) \citep{Lahlou2023} sidesteps Bayesian inference by training a secondary model $g(x)$ on held-out residuals to predict the primary model's expected error at each input. In generalized risk terms, recent theory decomposes total risk into Bayes risk (aleatoric) and excess risk (epistemic) using proper scoring rules and Bregman divergences \citep{KotelevskiiPanov2025}. Our implementation instantiates this idea empirically for ranking: total predicted rank displacement decomposes into an estimated noise floor $a(x)$ and excess model error $\ehat(x)$.

A parallel line of work questions how cleanly aleatoric and epistemic components can be disentangled in practice, especially under distribution shift \citep{Mucsanyi2024}. For deployment, we take an operational stance: what matters is whether $\ehat$ improves decisions (abstention, caps, risk limits), not whether it perfectly isolates a theoretical quantity.

\subsection{Uncertainty in portfolio construction}
Uncertainty-aware portfolio construction is commonly operationalized through multiplicative sizing rules, scaling exposure inversely with uncertainty \citep{Spears2021}. Related theory derives portfolios that shrink toward safety as parameter uncertainty increases \citep{Garlappi2007}. More recently, \citet{Liu2026} propose uncertainty-adjusted sorting, using uncertainty bounds to form long/short portfolios, and report improvements across multiple ML model families.

Our setting differs in a crucial way: we study cross-sectional \emph{ranking} models rather than return forecasts. In ranking, the strongest trade ideas live in the score tails, but those tails are also where rank displacement is largest, yielding a structural coupling between uncertainty and signal magnitude. This coupling can render standard inverse-uncertainty sizing counterproductive. A related idea is to residualize uncertainty on signal context before acting on it \citep{Hentschel2025}; our findings suggest that, in ranking, a discrete tail-risk cap can be more robust than continuous multiplicative rules.

Finally, learning from out-of-sample forecast errors improves portfolio outcomes in optimized settings \citep{BarrosoSaxena2021}, consistent with the general principle that systematic error learning is valuable---though that work does not address ranking geometry or the signal--uncertainty coupling we document.

\subsection{Selective prediction and abstention}
Selective prediction formalizes a reject option: the model may abstain on inputs where it anticipates high error \citep{ElYanivWiener2010,GeifmanElYaniv2017}. Modern treatments study coverage--risk tradeoffs and connections to conformal prediction \citep{ChaudhuriLopezPaz2023}. Extensions consider hybrid rules that reject the most uncertain predictions and apply confidence-weighted actions within the accepted set \citep{Franc2021,Rabanser2025}.

We implement abstention at two granularities simultaneously. At the strategy level, the regime-trust gate $\G(t)$ decides whether to deploy the ranker on a given date. At the position level, $\ehat$-Cap reduces exposure only for the most uncertain tail, a soft form of within-portfolio rejection. This separation is empirically motivated: cross-sectional uncertainty can rank stock-level error risk, but aggregated per-stock uncertainty measures are not reliable detectors of regime-wide model failure.

\subsection{Regime detection in quantitative finance}
Regime-switching models have a long history in econometrics and portfolio allocation \citep{Hamilton1989,Nystrup2017}, with extensions to robust optimization and modern learning-based approaches \citep{OprisorKwon2021,Pun2023}. These methods typically infer regimes from market observables (returns, volatility, correlations) and then condition portfolio choices on the inferred state.

Our approach is model-specific: the gate $\G(t)$ measures trailing realized efficacy of the model itself (using matured labels under a structural lag to preserve PIT safety), rather than market conditions alone. This distinction matters empirically: market stress proxies (e.g., VIX percentile) need not align with \emph{model failure}, motivating a direct ``does the model work?'' monitor.

\subsection{Positioning this paper}
Table~\ref{tab:positioning} summarizes the closest prior work and our contribution.

\begin{table}[H]
\centering
\small
\setlength{\tabcolsep}{6pt}
\renewcommand{\arraystretch}{1.2}
\begin{tabular}{@{}p{2.8cm}p{2.5cm}p{3.2cm}p{2.8cm}p{2.2cm}@{}}
\toprule
\textbf{Work} & \textbf{Task} & \textbf{UQ method} & \textbf{Action rule} & \textbf{Regime control} \\
\midrule
\citet{Spears2021}   & Futures return      & Bayesian NN posterior             & Multiplicative sizing        & Variance threshold \\
\citet{Liu2026}      & Return prediction   & Rolling residual quantiles        & Uncertainty-adj.\ sorting    & None \\
\citet{Garlappi2007} & Portfolio opt.\     & Ambiguity / model uncertainty     & Shrinkage toward safety      & None \\
\textbf{This paper}  & Cross-sec.\ ranking & DEUP on rank displacement         & Tail-risk cap ($\ehat$-Cap)  & Binary gate $\G(t)$ \\
\bottomrule
\end{tabular}
\caption{How this work relates to prior uncertainty-aware portfolio methods. In ranking, uncertainty can be structurally coupled to signal strength, motivating discrete tail-risk control plus a model-specific regime gate.}
\label{tab:positioning}
\end{table}

\FloatBarrier

\section{Problem Setup}

\label{sec:setup}

\subsection{Cross-Sectional Equity Ranking}
On each trading date $t$, we observe a dynamic universe of $N_t$ AI-exposed U.S.\ equities and a PIT-safe feature vector $x_{i,t}$ for each stock $i \in \{1,\dots,N_t\}$. A ranking model outputs a real-valued score $s_{i,t} \in \mathbb{R}$ (higher is better), inducing a cross-sectional ordering.

We evaluate ranking quality using the Spearman rank information coefficient (RankIC) between scores and subsequent realized \emph{excess} returns relative to a benchmark:
\begin{equation}
\RankIC(t;\tau)=\rho_S\!\left(\{s_{i,t}\}_{i=1}^{N_t},\;\{r^{(\tau)}_{i,t}\}_{i=1}^{N_t}\right),
\label{eq:rankic}
\end{equation}
where $r^{(\tau)}_{i,t}$ denotes the realized excess return of stock $i$ over the forward horizon $\tau \in \{20,60,90\}$ trading days (primary: $\tau=20$), and $\rho_S(\cdot,\cdot)$ is Spearman correlation computed cross-sectionally within each date.

\paragraph{Universe and sample.}
The sample spans February 2016 through February 2025, yielding 2{,}277 trading dates and 109 walk-forward evaluation folds. The benchmark is the Invesco QQQ ETF.

The investable universe is capped at 100 names by construction, but $N_t$ varies through time due to IPO entry and tradability filters. Measured directly from evaluation outputs (unique tickers per evaluation date, $\tau=20$d), the realized universe size distribution is:
\[
\text{ALL: mean }83.9,\ \text{median }83,\ p10=71,\ p90=98,\ \min=67,\ \max=99;
\]
\[
\text{DEV (}\le 2023\text{): mean }81.8,\ \text{median }80,\ p10=70,\ p90=97,\ \min=67,\ \max=98;
\]
\[
\text{FINAL (}\ge 2024\text{): mean }98.4,\ \text{median }98,\ p10=98,\ p90=99,\ \min=98,\ \max=99.
\]
The tightening in FINAL reflects that by 2024--2025 nearly all universe members are listed and pass tradability thresholds, whereas DEV includes pre-IPO periods for later entrants.

\subsection{Data, Tradability, and Return Definitions}
\label{sec:data-tradability}

\paragraph{Data sources and price conventions.}
Daily market data consist of OHLCV series used for feature engineering, portfolio simulation, and label construction. Prices are \emph{split-adjusted close} series (close-to-close, 4:00 PM ET) obtained from a vendor endpoint that returns split-adjusted prices directly. Corporate actions are treated as follows:
\begin{itemize}[leftmargin=*, itemsep=2pt]
  \item \textbf{Splits:} handled implicitly via vendor-provided split-adjusted price histories (no additional split adjustment is applied internally).
  \item \textbf{Dividends:} incorporated explicitly into labels via dividend yield computed from dividends with \emph{ex-date} in $(t,t+\tau]$, which is a conservative PIT-safe choice.
\end{itemize}

\paragraph{Benchmark-relative excess returns (total return).}
All targets and RankIC evaluations are performed on \emph{excess total returns} relative to QQQ. For horizon $\tau$ trading days, define stock total return as price return plus dividend yield:
\[
R_{i,t}^{(\tau)} \;=\; \left(\frac{P_{i,t+\tau}}{P_{i,t}} - 1\right) \;+\; \frac{\sum_{\text{ex-date}\in(t,t+\tau]} d_{i,\text{ex}}}{P_{i,t}},
\qquad
R_{\mathrm{QQQ},t}^{(\tau)} \;=\; \left(\frac{P_{\mathrm{QQQ},t+\tau}}{P_{\mathrm{QQQ},t}} - 1\right) \;+\; \frac{\sum_{\text{ex-date}\in(t,t+\tau]} d_{\mathrm{QQQ},\text{ex}}}{P_{\mathrm{QQQ},t}}.
\]
The realized excess return used throughout is then
\begin{equation}
r_{i,t}^{(\tau)} \;=\; R_{i,t}^{(\tau)} - R_{\mathrm{QQQ},t}^{(\tau)}.
\label{eq:excess-return}
\end{equation}
If benchmark dividend data are unavailable, labels fall back to stock total return minus benchmark \emph{price-only} return under a logged fallback policy.

\paragraph{Tradability filters and investable universe construction.}
Universe membership is defined \emph{as-of} each rebalance date $t$ by applying a deterministic tradability-and-relevance waterfall, then taking the top 100 by market capitalization:
\begin{itemize}[leftmargin=*, itemsep=2pt]
  \item \textbf{Exchange/type filter:} U.S.\ common stocks only (Polygon ticker type ``CS'').
  \item \textbf{Minimum average price:} $P_{i,t} \ge \$5.00$.
  \item \textbf{Minimum liquidity (ADV):} average daily \emph{dollar} volume $\mathrm{ADV}_{i,t} \ge \$1{,}000{,}000$,
  computed as $\mathrm{ADV}_{i,t} = \left(\frac{1}{20}\sum_{j=1}^{20} \mathrm{Vol}_{i,t-j}\right)\cdot P_{i,t}$ using the most recent 20 trading days of volume history.
  \item \textbf{AI exposure filter:} membership in a fixed AI-themed ticker list (100 unique tickers across categories), applied as an exact ticker match.
  \item \textbf{Max constituents:} select the top 100 by market cap after the above filters.
\end{itemize}
No minimum market-cap threshold is imposed by default beyond the top-100 cap.

\paragraph{Survivorship, delistings, and identifier stability.}
Universe construction is survivorship-aware when Polygon is used: constituents are queried \emph{as-of} date $t$ with \texttt{active\_only=False}, allowing historically active but subsequently delisted tickers to be included in past universes when data exist. A security master tracks stable identifiers across ticker changes (e.g., FB$\rightarrow$META) and records security events (delisting, merger, spinoff, bankruptcy), enabling timestamped membership replay rather than retroactive dropping.

\paragraph{Label availability and missing-data policy.}
Labels require entry/exit prices for both the stock and benchmark at $(t,t+\tau)$. If any of \{entry price, exit price, benchmark entry, benchmark exit\} is missing, the label is not formed and the corresponding row is dropped from training/evaluation for that horizon.
Across all horizons and dates, 600{,}855 out of 688{,}800 candidate label attempts are valid (87.2\% coverage; 12.8\% dropped). The dropped set is dominated by correct pre-IPO behavior for later entrants (approximately 98\% of dropped rows), with a single mid-sample exception attributable to an ADR leaving the U.S.\ market (ABB), which accounts for essentially all non-pre-IPO gaps. Within the realized 20d evaluation window, label availability is effectively complete: 99.95\% of $(\text{ticker}\times\text{date})$ pairs have a valid 20d label, with the remaining 0.05\% corresponding to the ABB exit episode.

\paragraph{Important limitation: delisting returns.}
CRSP-style delisting returns are not modeled. If a security delists within the holding period and no exit price is available at $t+\tau$, the label is dropped rather than replaced with a terminal delisting return. In this 100-name AI universe this behavior is empirically rare; for broader universes, explicit delisting-return modeling would be required to avoid bias.

\paragraph{Feature missingness and preprocessing.}
Missingness is handled component-wise:
\begin{itemize}[leftmargin=*, itemsep=2pt]
  \item \textbf{Base ranker features:} LightGBM is trained on raw PIT-safe features with native NaN handling (no explicit imputation).
  \item \textbf{Labels:} rows with missing excess returns are dropped prior to training/evaluation.
  \item \textbf{DEUP meta-model features:} uncertainty-model features are zero-filled (\texttt{fillna(0)}) in the DEUP pipeline.
\end{itemize}
No winsorization of the excess-return labels is applied in the frozen LightGBM baseline pipeline.

\subsection{Base Ranker (LightGBM)}
\label{sec:base-model}
The primary ranker is a LightGBM gradient-boosted tree ensemble trained on seven PIT-safe features: three momentum signals (\texttt{mom\_1m}, \texttt{mom\_3m}, \texttt{mom\_12m}), two realized-volatility measures (\texttt{vol\_20d}, \texttt{vol\_60d}), an average daily volume proxy (\texttt{adv\_20d}), and the stock's cross-sectional rank on the current date (\texttt{cross\_sectional\_rank}). The training target is the stock's excess return versus QQQ at the specified horizon.

\paragraph{Walk-forward training.}
Models are trained in an expanding-window walk-forward configuration. For evaluation fold $k \in \{1,\dots,109\}$, training uses all data from folds $1$ through $k-1$ with a minimum 90-trading-day embargo between the most recent training label-maturation date and the earliest prediction date in fold $k$. Overlapping labels are purged to prevent leakage from forward-return overlap.

\subsection{Shadow Portfolio (Economic Signal Test)}
\label{sec:shadow-portfolio}
To translate cross-sectional signal quality into an economic sanity check, we construct a simple shadow long--short portfolio. On each rebalance date (non-overlapping 20-trading-day intervals), the top-$K$ stocks by score form the long leg and the bottom-$K$ form the short leg, with $K=10$ and equal weights within each leg. For reporting, we form a non-overlapping monthly return series by selecting the first trading day of each calendar month and using the corresponding 20-trading-day forward return; this is the common return series used for Sharpe and Crisis MaxDD in all deployment-policy tables unless explicitly stated otherwise. A transaction cost of 10 basis points per rebalance is applied. This portfolio is deliberately simple---fixed $K$, no optimization---so that performance primarily reflects ranking quality rather than portfolio-construction sophistication.

We report both raw and volatility-sized variants of this shadow portfolio. Over the full walk-forward sample (ALL), the volatility-sized 20d LightGBM shadow portfolio achieves Sharpe 2.73 with 82.6\% monthly hit rate and $-18.1\%$ maximum drawdown. Performance degrades in the FINAL holdout (2024 onward): Sharpe 1.91 with 71.4\% hit rate, consistent with a genuine regime shift in the AI-exposed equity universe.

\subsection{Evaluation Protocol (DEV/FINAL)}
\label{sec:eval-protocol}
All results are reported under a strict DEV/FINAL protocol:
\begin{itemize}[leftmargin=*, itemsep=2pt]
    \item \textbf{DEV period} (2016--2023, 95 months): all hyperparameter tuning, calibration, threshold selection, and model iteration use DEV only.
    \item \textbf{FINAL period} (2024 onward, 14 months): evaluated exactly once with all parameters frozen from DEV (no re-tuning).
\end{itemize}
The walk-forward structure ensures each prediction is out-of-sample relative to its fold-specific training set. The DEV/FINAL split adds a second, end-to-end temporal holdout to reduce implicit overfitting through repeated experimentation.

\subsection{Loss Definition for Uncertainty: Rank Displacement}
\label{sec:loss-definition}
Our uncertainty framework requires a per-prediction loss aligned with ranking. We use \emph{rank displacement}:
\begin{equation}
\ell(x_{i,t};\tau) \;=\;
\left|
\operatorname{rank}_{\%}\!\left(r^{(\tau)}_{i,t}\right)
-
\operatorname{rank}_{\%}\!\left(s_{i,t}\right)
\right|,
\label{eq:rank-loss}
\end{equation}
where $\operatorname{rank}_{\%}(\cdot)$ denotes the cross-sectional percentile rank on date $t$. Rank displacement lies in $[0,1]$ and directly measures ranking error: a stock scored in the 90th percentile that realizes returns at the 40th percentile has $\ell=0.50$.

This loss is preferable to return-space errors (e.g., MAE) for two practical reasons. First, it directly targets the object used in evaluation (cross-sectional ordering) and is therefore tightly aligned with RankIC. Second, it avoids conflating ranking failure with idiosyncratic volatility: in our experiments, rank displacement has near-zero correlation with stock-level realized volatility at 20d ($\rho=0.054$), supporting the use of rank displacement as an error target for uncertainty modeling rather than a proxy for volatility.


\section{Method}
\label{sec:method}

We present a two-level uncertainty architecture in four parts: (1)~adaptation of DEUP to cross-sectional ranking loss, (2)~the structural coupling diagnostic that motivates the architecture, (3)~the regime-trust gate and position-level tail-risk cap, and (4)~the deployment policy variants tested.

\subsection{DEUP for Ranking Loss}
\label{sec:deup-ranking}

\subsubsection{Error Predictor $g(x)$}

Following \citet{Lahlou2023}, we train a secondary model $g(x)$ to predict the primary model's expected rank displacement at each input. Concretely, let $\ell_{i,t}$ denote the realized rank displacement (Equation~\ref{eq:rank-loss}) of stock $i$ on date $t$. The error predictor is a LightGBM regression model:
\begin{equation}
g(x_{i,t}) \approx \mathbb{E}[\ell_{i,t} \mid x_{i,t}],
\label{eq:gx}
\end{equation}
trained walk-forward on held-out residuals from the primary model. The walk-forward protocol mirrors the base model: an expanding window over 109 folds, with predictions generated for folds 21--109 (requiring a minimum of 20 training folds, yielding 89 prediction folds and 161{,}863 predictions per horizon).

The error predictor uses 11 features drawn from three categories:

\begin{table}[h]
\centering
\small
\caption{Features used in the error predictor $g(x)$.}
\label{tab:gx-features}
\begin{tabular}{lll}
\toprule
\textbf{Category} & \textbf{Feature} & \textbf{Purpose} \\
\midrule
Per-prediction & \texttt{score}, \texttt{abs\_score} & Prediction magnitude/strength \\
               & \texttt{cross\_sectional\_rank} & Position in today's ranking \\
\midrule
Stock-level    & \texttt{vol\_20d}, \texttt{vol\_60d} & Realized volatility \\
               & \texttt{mom\_1m} & Recent momentum \\
               & \texttt{adv\_20d} & Liquidity proxy \\
\midrule
Market regime  & \texttt{vix\_percentile\_252d} & Rolling VIX percentile \\
               & \texttt{market\_regime\_enc} & Bull/Neutral/Bear encoding \\
               & \texttt{market\_vol\_21d} & Market-level volatility \\
               & \texttt{market\_return\_21d} & Recent market return \\
\bottomrule
\end{tabular}
\end{table}

The LightGBM hyperparameters are deliberately conservative to reduce meta-overfitting: $n_{\text{estimators}} = 50$, $\text{max\_depth} = 3$, $\text{num\_leaves} = 8$, $\text{min\_child\_samples} = 50$, $\text{learning\_rate} = 0.05$, with subsampling at 80\% of rows and columns.

At the 20d horizon, $g(x)$ achieves Spearman $\rho(g,\ell)=0.192$ with realized rank displacement in walk-forward evaluation. Feature importance analysis reveals that \texttt{cross\_sectional\_rank} is the dominant predictor at all horizons (split importance 101--120), followed by \texttt{score} and \texttt{vol\_60d}. This dominance is economically meaningful: stocks at the extremes of the model's cross-sectional ranking have systematically different---and predictable---error profiles than mid-ranked stocks.

\subsubsection{Aleatoric Baseline $a(\cdot)$}
\label{sec:aleatoric}

The aleatoric component $a(\cdot)$ estimates irreducible ranking noise: the floor of rank displacement that persists even for an optimal predictor under the prevailing information set. In the \citet{KotelevskiiPanov2025} framework, this corresponds to the Bayes-risk component of total predictive risk.

\paragraph{Key point (narrow claim): within-date ordering vs.\ deployment magnitudes.}
For cross-sectional stock ranking, our primary diagnostic use of DEUP is the \emph{within-date} ordering of failure risk. For any fixed date $t$, if the aleatoric term is constant across stocks (i.e., $a(x_{i,t}) \equiv a(t)$), then the cross-sectional ordering of
\[
\hat{e}(x_{i,t})=\max\!\bigl(0,\,g(x_{i,t})-a(x_{i,t})\bigr)
\]
is determined entirely by $g(x_{i,t})$: subtracting a date-level constant does not change the rank order across $i$. This argument justifies using a simple per-date floor for \emph{ranking diagnostics} (e.g., quintile monotonicity, within-date $\rho(\hat{e},\ell)$). It does \emph{not} justify using a hindsight floor for \emph{deployment actions} that depend on the \emph{magnitude} of $\hat{e}$ (e.g., caps, conformal scaling), where PIT-safety is required.

\paragraph{Candidate estimators.}
We evaluate a tiered set of increasingly structured proxies:

\begin{enumerate}
    \item \textbf{Tier~0 (Inverse IQR):} $a(t) = c / (\mathrm{IQR}(\mathbf{r}_t) + \varepsilon)$, calibrated so that $\mathrm{median}(a) \approx \mathrm{median}(\ell)$. \textit{Killed} at all horizons ($\rho \approx 0$ with mean rank loss). Cross-sectional return dispersion conflates signal dispersion with noise dispersion and does not identify hard-to-rank days.
    \item \textbf{Tier~1 (Factor-Residual IQR):} regress cross-sectional returns on sector dummies and momentum, then apply Tier~0 to residuals. Also \textit{killed} ($\rho \approx 0$): removing factor structure does not resolve the signal-vs-noise ambiguity.
    \item \textbf{Tier~2 (Per-Stock LGB Quantile Regression):} walk-forward LightGBM quantile models predicting the 25th and 75th percentiles of stock-level rank displacement from characteristics (\texttt{vol\_20d}, \texttt{adv\_20d}, \texttt{market\_vol\_21d}, \texttt{vix\_percentile\_252d}, \texttt{mom\_1m}, \texttt{sector\_enc}). \textit{Passes} at 60d ($\rho = 0.317$, consistent across three base models); \textit{killed} at 20d ($\rho = 0.024$).
\end{enumerate}

The failure of Tiers~0 and~1 is itself a finding: cross-sectional dispersion of realized returns---the most natural proxy for ranking ``noise''---has no predictive power for ranking difficulty in this setting. At 20d, ranking noise is dominated by short-horizon idiosyncratic movements with extremely low autocorrelation ($\rho_{\text{lag-20}} = 0.061$), rendering it effectively unpredictable from observable regime proxies.

\paragraph{(A) Oracle / hindsight floor $a_{\mathrm{oracle}}(t)$.}
For diagnostic analysis only, we define an oracle per-date floor using realized same-date rank losses:
\begin{equation}
a_{\mathrm{oracle}}(t) \;=\; P_{10}\!\left(\boldsymbol{\ell}_t\right),
\label{eq:aleatoric-oracle}
\end{equation}
where $\boldsymbol{\ell}_t=\{\ell_{i,t}\}_{i=1}^{N_t}$ is the cross-section of realized rank displacements on date $t$. This stabilizes the scale of $\hat{e}$ for visualization and diagnostics, while leaving the \emph{within-date} stock ordering entirely determined by $g(x)$.

\paragraph{(B) Deployable PIT-safe floor $a_{\mathrm{PIT}}(t)$.}
All deployment-facing uses of $\hat{e}$ that depend on its \emph{magnitude} (caps and conformal scaling) use a PIT-safe aleatoric baseline computed only from \emph{matured} historical losses, aligned with the horizon lag $\tau$.

\textit{Primary PIT-safe definition (rolling matured $P_{10}$).} Let $\tau$ denote the horizon in trading days (e.g., $\tau=20$ at 20d). For a trailing window length $W$ (we use $W=60$ operationally; we also test $W=252$ as a robustness alternative), define:
\begin{equation}
a_{\mathrm{PIT}}(t)
\;=\;
P_{10}\!\left(
\left\{\ell_{i,u}:\; u \in [t-\tau-W,\; t-\tau]\right\}
\right),
\qquad W\in\{60,252\}.
\label{eq:aleatoric-pit}
\end{equation}
This uses only losses whose forward returns have fully matured by time $t$.

\textit{Robustness alternative (expanding-window constant baseline).} As a simpler, lower-variance deployable proxy, we also consider:
\begin{equation}
a_{\mathrm{EXP}}(t)
\;=\;
\mathrm{median}_{u \le t-\tau}\;
P_{10}\!\left(\boldsymbol{\ell}_u\right),
\label{eq:aleatoric-exp}
\end{equation}
which is constant within a given date and updated only using matured history.

\paragraph{Evaluation protocol.}
Unless explicitly labeled as oracle/diagnostic, all reported deployment-policy results that use $\hat{e}$ magnitudes (caps and conformal normalization) are computed with $a_{\mathrm{PIT}}(t)$ (Equation~\ref{eq:aleatoric-pit}). We additionally report a hindsight-vs-deployable ablation by comparing results under $a_{\mathrm{oracle}}(t)$ versus $a_{\mathrm{PIT}}(t)$ (and $a_{\mathrm{EXP}}(t)$), quantifying sensitivity to the aleatoric baseline.

\subsubsection{Epistemic Signal $\hat{e}(x)$}
\label{sec:ehat}

The epistemic uncertainty signal is defined as predicted rank displacement beyond the irreducible noise floor:
\begin{equation}
\hat{e}(x_{i,t}) = \max\bigl(0,\; g(x_{i,t}) - a(x_{i,t})\bigr),
\label{eq:ehat}
\end{equation}
where the $\max(0,\cdot)$ operator enforces non-negativity.

\paragraph{Oracle vs.\ deployable variants.}
Because $a(\cdot)$ admits both oracle and PIT-safe constructions, we define two corresponding epistemic signals:
\begin{align}
\hat{e}_{\mathrm{oracle}}(x_{i,t})
&=
\max\!\left(0,\; g(x_{i,t}) - a_{\mathrm{oracle}}(t)\right),
\label{eq:ehat-oracle}
\\
\hat{e}_{\mathrm{PIT}}(x_{i,t})
&=
\max\!\left(0,\; g(x_{i,t}) - a_{\mathrm{PIT}}(t)\right).
\label{eq:ehat-pit}
\end{align}
The within-date \emph{ordering} of $\hat{e}_{\mathrm{oracle}}$ and $\hat{e}_{\mathrm{PIT}}$ is identical (both are date-level shifts of $g(x)$), but their \emph{magnitudes} differ and therefore yield different cap/scaling behavior. Unless noted otherwise, all \emph{deployable policy experiments} in this paper use $\hat{e}_{\mathrm{PIT}}$.

\paragraph{Validation.} We validate $\hat{e}(x)$ through six diagnostic tests:

\begin{enumerate}[label=\textbf{(\Alph*)}]
    \item \textbf{Disentanglement (Residualization):} After regressing $\hat{e}$ on \texttt{vol\_20d}, \texttt{vol\_60d}, \texttt{vix\_percentile\_252d}, and \texttt{mom\_1m} via OLS and taking residuals, the residualized $\hat{e}$ retains $\rho = 0.11$--$0.24$ with rank loss across all horizons. The maximum absolute correlation between $\hat{e}$ and VIX is 0.074. \textit{Pass:} $\hat{e}$ is not repackaged volatility.
    
    \item \textbf{Quintile Monotonicity:} Stocks sorted into quintiles by $\hat{e}$ exhibit strictly monotonically increasing mean rank loss at 20d and 90d in both DEV and FINAL. Spearman $\rho = 1.0$ across all four conditions (DEV/FINAL $\times$ 20d/90d). In the FINAL holdout, the Q5/Q1 rank-loss ratio is 1.69 at 20d and 1.88 at 90d---\emph{stronger} separation than in DEV (1.51, 1.53), the opposite of overfitting.
    
    \item \textbf{AUROC for High-Loss Events:} Stock-level AUROC for above-median rank loss ranges from 0.53 to 0.61. The signal identifies individual high-error predictions, though with moderate discriminative power.
    
    \item \textbf{Regime Failure Detection:} Per-stock $\hat{e}$ does \emph{not} spike during the 2024 regime failure. Crisis-period mean $\hat{e}$ (0.259) is statistically indistinguishable from non-crisis (0.271). \textit{Expected:} $\hat{e}$ is a per-stock signal by design; regime failure affects all stocks uniformly and requires a separate per-date detector (Section~\ref{sec:regime-gate}).
    
    \item \textbf{Baseline Dominance:} $\hat{e}$ and $g(x)$ achieve 3--10$\times$ higher $\rho$ with rank loss than \texttt{vol\_20d} ($\rho = 0.01$--$0.05$) or VIX percentile ($\rho = -0.02$ to $0.05$). In the FINAL holdout, volatility and VIX lose predictive power entirely ($\rho \approx 0$); $\hat{e}$ maintains or improves its correlation. See Table~\ref{tab:baseline-comparison}.
    
    \item \textbf{Feature Importance:} Per-prediction features (\texttt{cross\_sectional\_rank}, \texttt{score}) account for 39--56\% of $g(x)$'s split importance, confirming that $g(x)$ captures per-stock prediction uncertainty rather than market-level regime information.
\end{enumerate}

\begin{table}[t]
\centering
\small
\caption{Per-stock uncertainty signal comparison: Spearman $\rho$ with realized rank loss. $\hat{e}$ and $g(x)$ dominate heuristic baselines by $3$--$10\times$ and maintain predictive power in the FINAL holdout where volatility-based signals collapse.}
\label{tab:baseline-comparison}
\begin{tabular}{lcccc}
\toprule
\textbf{Signal} & \textbf{20d ALL} & \textbf{20d FINAL} & \textbf{90d ALL} & \textbf{90d FINAL} \\
\midrule
$\hat{e}(x)$ (DEUP)  & 0.144 & \textbf{0.192} & 0.146 & \textbf{0.248} \\
$g(x)$ (raw)          & \textbf{0.192} & 0.218 & 0.161 & 0.262 \\
\texttt{vol\_20d}     & 0.047 & 0.010 & 0.035 & 0.009 \\
VIX percentile        & 0.018 & $-$0.022 & 0.053 & $-$0.020 \\
$|\text{score}|$      & 0.096 & 0.065 & 0.018 & 0.013 \\
\bottomrule
\end{tabular}
\end{table}

\paragraph{Stability.} The signal is positive under all tested conditioning cuts: pre-2020 vs.\ post-2020, low-VIX vs.\ high-VIX regimes, and low-volatility vs.\ high-volatility stocks ($\rho = 0.03$--$0.21$ across all conditions at all horizons).

\paragraph{Robustness across base models.} We replicate the DEUP pipeline on a second base model---Rank Average~2, an ensemble of LightGBM and FinText-TSFM signals---to verify that the uncertainty signal is not an artifact of the LightGBM architecture. RA2 produces $\hat{e}$ with 35\% higher $\rho(\hat{e}, \ell)$ at 20d (0.194 vs.\ 0.144) and 44\% higher at 60d (0.153 vs.\ 0.106), confirming that DEUP uncertainty quality responds to base-model robustness. The structural coupling between $\hat{e}$ and $|\text{score}|$ persists for RA2 (Section~\ref{sec:structural-coupling}), confirming it is a model-agnostic property of cross-sectional ranking rather than an LGB-specific artifact.


\subsection{The Structural Coupling Problem}
\label{sec:structural-coupling}

The standard prescription for deploying uncertainty estimates in portfolio construction is inverse-uncertainty sizing: allocate less capital to predictions with higher uncertainty. In return-prediction settings, this is well-motivated and empirically successful~\citep{Liu2026, Spears2021}. We show that it fails structurally in cross-sectional ranking due to a coupling between epistemic uncertainty and signal strength.

\subsubsection{Empirical Measurement}

On each trading date $t$, we compute the cross-sectional Spearman correlation between the epistemic signal and the absolute model score:
\begin{equation}
\rho_t = \rho_S\bigl(\hat{e}_t,\; |\mathbf{s}_t|\bigr).
\label{eq:coupling}
\end{equation}
Across 1{,}865 trading dates (the walk-forward prediction window), the median cross-sectional correlation is $\rho = 0.616$, with positive correlation on more than 90\% of dates.

\subsubsection{Mechanistic Explanation}

The coupling arises from the interaction between the ranking loss target and the error predictor's feature structure:

\begin{enumerate}
    \item \textbf{Extreme ranks have higher rank displacement by construction.} A stock scored at the 95th percentile cross-sectionally has room to ``fall'' by up to 95 percentile points; a stock at the 50th percentile can fall by at most 50. Expected rank displacement scales with distance from the median rank.
    
    \item \textbf{\texttt{cross\_sectional\_rank} is $g(x)$'s dominant feature.} Feature importance analysis (Diagnostic~F) shows that \texttt{cross\_sectional\_rank} is the single most important predictor of rank displacement at all horizons (split importance 101--120 out of 300--500 total), followed by \texttt{score} itself.
    
    \item \textbf{Extreme ranks carry the strongest trading signal.} The shadow portfolio is constructed from the top-10 and bottom-10 stocks---precisely the extreme ranks where $\hat{e}$ is highest.
\end{enumerate}

The consequence is that inverse-uncertainty sizing $w_i \propto 1/\sqrt{\hat{e}_i + \varepsilon}$ systematically reduces weight on the stocks that contribute most to the long--short spread. This is not a failure of DEUP as an uncertainty estimator---$\hat{e}$ genuinely predicts error magnitude---but a fundamental incompatibility between the ranking-loss uncertainty metric and the multiplicative sizing paradigm. In return-prediction settings~\citep{Liu2026, Spears2021}, the forecast magnitude and forecast uncertainty are not structurally constrained to covariate; in ranking settings, they are.

\subsubsection{Empirical Consequences}

Table~\ref{tab:sizing-failure} confirms the economic impact. In the gated shadow portfolio (binary gate at $G(t) \geq 0.2$), inverse-$\hat{e}$ sizing (Variant~4: residualized-$\hat{e}$) and uncertainty-adjusted sorting (Variant~3: Liu et al.\ UA Sort with $\lambda = 0.05$, calibrated on DEV) both underperform simple volatility-based sizing, despite using a more informative uncertainty signal.

\begin{table}[t]
\centering
\small
\caption{Deployment-policy ablation (20d). Sharpe $\times\sqrt{12}$ from monthly non-overlapping returns. FINAL $N=14$ months.}
\label{tab:sizing-failure}
\begin{tabular}{lcccc}
\toprule
\textbf{Variant} & \textbf{ALL} & \textbf{DEV} & \textbf{FINAL} & \textbf{Crisis MaxDD} \\
\midrule
Ungated Raw (LGB) & 2.730 & 3.107 & 1.650 & -7.4\% \\
\midrule
1. Gate + Raw & 1.928 & 2.039 & -0.322 & -7.4\% \\
2. Gate + Vol & 1.886 & 1.971 & 0.375 & -8.4\% \\
4. Gate + Resid-$\hat{e}$ & 1.987 & 2.041 & 0.953 & -6.0\% \\
\textbf{6. Gate + Vol + $\hat{e}$-Cap} & \textbf{1.877} & \textbf{1.928} & \textbf{0.925} & \textbf{-6.7\%} \\
\bottomrule
\end{tabular}
\end{table}

\citet{Liu2026}'s UA Sort, which achieves substantial Sharpe improvements in their return-prediction setting, produces an optimal $\lambda$ of only 0.05 in our ranking system (calibrated on DEV via grid search)---effectively zero. The uncertainty adjustment is absorbed by the existing score structure because the information content of $\hat{e}$ is already embedded in the extreme-rank positions.


\subsection{Two-Level Deployment Architecture}
\label{sec:two-level}

The structural coupling motivates an architectural separation: rather than using uncertainty for continuous position sizing, we deploy it at two discrete levels that address qualitatively distinct failure modes.

\subsubsection{Level 1: Strategy-Level Regime Gate $G(t)$}
\label{sec:regime-gate}

\paragraph{Motivation.} Diagnostic~D establishes that per-stock $\hat{e}(x)$ cannot detect regime-level model failure: aggregated per-stock uncertainty (median, 90th percentile, cross-sectional spread of $\hat{e}$) achieves AUROC~$\approx 0.50$ for predicting whether the model will have a positive-RankIC day---no better than random. This is by design: $\hat{e}$ measures cross-sectional variation in prediction quality, while regime failure is a latent variable that affects all stocks uniformly. A separate per-date signal is required.

\paragraph{Binary target and evaluation alignment.}
We define a \emph{good day} as
\[
\text{good\_day}(t) = \mathbb{1}\!\left[\text{matured\_RankIC}(t) > 0\right],
\]
where \text{matured\_RankIC} at date $t$ is computed from a PIT-safe stream of realized RankIC values whose underlying $\tau$-day forward returns have fully matured (a structural $\tau$-day lag at horizon $\tau$). All candidate regime predictors are date-level aggregates (mean across stocks on each date) aligned by as-of date. To keep the AUROC direction consistent (higher predictor $\Rightarrow$ higher probability of a good day), we apply the same sign conventions as the implementation: $1-\text{VIX percentile}$, $-\text{market\_vol}_{21d}$, and $-\text{mean\_stock\_vol}_{20d}$. Any residual missing values are filled with zero prior to \texttt{roc\_auc\_score}. Early warm-up dates without a defined health signal are dropped (59 dates in ALL; none in FINAL).

\paragraph{Construction.} The regime-trust gate $G(t)$ is derived from a health index $H(t)$ that combines three PIT-safe signals:

\textit{(i)~Realized efficacy} $H_{\text{real}}(t)$: an exponentially weighted moving average (halflife~$= 30$, $\text{min\_periods} = 20$) of \emph{matured} daily RankIC values. At date $t$, only RankIC values from date $t - \tau$ (i.e., predictions whose returns have fully matured) are used, imposing a structural $\tau$-day lag (20 trading days at the 20d horizon). This is the only fully honest realized-efficacy measure---no future information contaminates the signal.

\textit{(ii)~Feature and score drift} $H_{\text{drift}}(t)$: a real-time (zero-lag) composite of three sub-signals: (a)~mean absolute z-score of key features (\texttt{vol\_20d}, \texttt{mom\_1m}, \texttt{adv\_20d}, \texttt{vix\_percentile\_252d}, \texttt{market\_vol\_21d}, \texttt{vol\_60d}) relative to their trailing 252-day means, (b)~the Kolmogorov--Smirnov statistic of today's score distribution versus a trailing 60-day reference, and (c)~mean pairwise return correlation over the trailing 20 days (a proxy for cross-sectional separability). Combined as $H_{\text{drift}} = 0.4 \cdot \text{feat\_drift} + 0.3 \cdot \text{score\_drift} + 0.3 \cdot \text{corr\_spike}$.

\textit{(iii)~Cross-expert disagreement} $H_{\text{disagree}}(t)$: the Spearman rank correlation between the primary model (LGB) and an alternative model (Rank Average~2) on the same date. Low correlation signals potential regime boundaries. If only one model is available, score dispersion relative to its expanding mean serves as a fallback proxy.

These are combined via expanding z-scores:
\begin{align}
H_{\text{raw}}(t) &= z_{\text{real}}(t) - \alpha \cdot z_{\text{drift}}(t) - \beta \cdot z_{\text{disagree}}(t), \quad \alpha = \beta = 0.3, \label{eq:H-raw} \\
H(t) &= \sigma\bigl(H_{\text{raw}}(t)\bigr) \in [0, 1], \label{eq:H-sigmoid} \\
G(t) &= \mathrm{clip}\!\left(\frac{H(t) - 0.3}{0.7 - 0.3},\; 0,\; 1\right), \label{eq:G-gate}
\end{align}
where $\sigma(\cdot)$ is the logistic sigmoid and $\mathrm{clip}(\cdot, 0, 1)$ clamps the output to $[0, 1]$. $H_{\text{real}}$ is the dominant component by design; drift and disagreement provide supplementary adjustment ($H_{\text{real}}$-only AUROC $= 0.715$ vs.\ combined $H(t)$ AUROC $= 0.721$).

\paragraph{Deployment rule.} $G(t)$ is applied as a \emph{binary} gate:
\begin{equation}
\text{Active}(t) = \mathbb{1}[G(t) \geq 0.2].
\label{eq:gate-rule}
\end{equation}
If inactive, the strategy holds zero exposure (cash). The threshold $\theta = 0.2$ is selected on DEV to maximize precision at a reasonable abstention rate. Binary abstention outperforms continuous scaling ($w(t) \propto G(t)$) because continuous throttling destroys recovery convexity: the strategy stays partially ``off'' during rebounds following regime failures, compressing realized Sharpe. This finding is consistent with \citet{Nystrup2017}'s observation that excessively frequent regime-switching hurts portfolio outcomes.

\paragraph{Regime-gate classifier performance.} Table~\ref{tab:regime-gate} summarizes the regime-trust classification results. In the FINAL holdout, common volatility proxies are weak or unstable: VIX percentile is effectively random (AUROC $\approx 0.50$) and mean stock volatility reverses sign (AUROC $<0.5$), while $H(t)$ and $G(t)$ retain strong discrimination.

\begin{table}[t]
\centering
\small
\caption{Regime-trust classification at 20d. $H(t)$ and $G(t)$ substantially outperform market-regime proxies. In the FINAL holdout, VIX percentile is effectively uninformative (AUROC $\approx 0.5$) and mean stock volatility becomes negatively informative (AUROC $<0.5$).}
\label{tab:regime-gate}
\begin{tabular}{lcc}
\toprule
\textbf{Predictor} & \textbf{AUROC (ALL)} & \textbf{AUROC (FINAL)} \\
\midrule
$H(t)$ (combined)        & 0.721 & \textbf{0.750} \\
$G(t)$ (gate)            & 0.710 & 0.743 \\
$H_{\text{real}}$-only   & 0.715 & --- \\
VIX percentile           & 0.449 & 0.504 \\
Market volatility        & 0.596 & 0.569 \\
Mean stock volatility    & 0.590 & 0.460 \\
\bottomrule
\end{tabular}
\end{table}

At the $G(t) \geq 0.2$ operating point, the gate achieves 80\% precision (when the system decides to trade, 80\% of those days have positive matured RankIC), 64\% recall, and 47\% abstention rate. The confusion matrix over the full walk-forward sample yields 937 true positives and 234 false positives out of 1{,}171 active days, with 520 true negatives and 527 false negatives. The $G(t)$ bucket analysis exhibits perfect monotonicity (Spearman $\rho = 1.0$): the lowest-$G$ bucket has mean RankIC~$= -0.011$ with 51\% bad days, while the highest-$G$ bucket has mean RankIC~$= +0.153$ with only 12\% bad days.

\paragraph{Multi-crisis validation.} To confirm that $G(t)$ is not overfit to the 2024 episode, we evaluate its behavior across five distinct stress episodes and three calm reference periods spanning 2016--2025, using only pre-existing frozen outputs (no retraining). We compare $G(t)$ against a VIX-percentile-based gate (abstain if VIX exceeds the 67th rolling-252-day percentile on $>50\%$ of window days). Results appear in Section~\ref{sec:multi-crisis-results}.

\subsubsection{Level 2: Position-Level $\hat{e}$-Cap (Tail-Risk Guard)}
\label{sec:ehat-cap}

Within active trading days (when $G(t) \geq 0.2$), the structural coupling prevents $\hat{e}$ from being used as a sizing denominator. However, $\hat{e}$ retains value as a \emph{discrete} tail-risk guard. We implement a percentile-based cap using the \emph{deployable} epistemic signal $\hat{e}_{\mathrm{PIT}}$:
\begin{equation}
w_i = \begin{cases}
w_i^{\text{vol}} \cdot \kappa & \text{if } \hat{e}_{\mathrm{PIT}}(x_{i,t}) > P_p(\hat{e}_{\mathrm{PIT},t}), \\
w_i^{\text{vol}} & \text{otherwise},
\end{cases}
\label{eq:ehat-cap}
\end{equation}
where $w_i^{\text{vol}}$ is the volatility-sized weight, $p$ is the cap percentile (85th, calibrated on DEV), and $\kappa = 0.70$ is the cap weight (i.e., a 30\% weight reduction). This caps the top 15\% most uncertain stocks by $\hat{e}_{\mathrm{PIT}}$ on each date, leaving the remaining 85\% untouched.

\paragraph{Why PIT-safe $a(t)$ is required for the cap.}
The cap threshold is cross-sectional and therefore sensitive to shifts and kinks in the distribution of $\hat{e}$ across dates. Any look-ahead in the aleatoric floor would leak into the \emph{magnitude} of $\hat{e}$ and hence into whether a name falls above the $P_p(\cdot)$ cutoff. For this reason, all cap results use $\hat{e}_{\mathrm{PIT}}$ (Equation~\ref{eq:ehat-pit}), i.e., a PIT-safe $a_{\mathrm{PIT}}(t)$.

The $\hat{e}$-cap operates as a selective-prediction mechanism within the accepted portfolio~\citep{ChaudhuriLopezPaz2023}: predictions in the extreme tail of predicted epistemic uncertainty receive reduced capital allocation, implementing a form of ``soft abstention'' at the position level. Crucially, this avoids the structural coupling problem because it does not continuously modulate all weights by $\hat{e}$---it applies a discrete filter only to the tail, where the coupling between $\hat{e}$ and signal strength is most extreme.

\subsection{The Combined System}
\label{sec:combined-system}

The recommended deployment system combines three layers:
\begin{enumerate}
    \item \textbf{Strategy gate} (Section~\ref{sec:regime-gate}): If $G(t) < 0.2$, hold cash. No exposure.
    \item \textbf{Volatility sizing:} When active, compute $w_i^{\text{vol}} = s_{i,t} \cdot \min\bigl(1,\; c_{\text{vol}} / \sqrt{\texttt{vol\_20d}_i + \varepsilon}\bigr)$, where $c_{\text{vol}}$ is calibrated on DEV such that the median weight $\approx 0.7$. Select top-10/bottom-10 by sized score for the long/short legs.
    \item \textbf{Tail-risk cap} (Section~\ref{sec:ehat-cap}): Within the selected portfolio, reduce weight by 30\% for stocks above the 85th percentile of $\hat{e}_{\mathrm{PIT}}$ on that date.
\end{enumerate}

This architecture separates two qualitatively distinct risk-management questions:

\begin{table}[h]
\centering
\small
\caption{The two-level deployment architecture addresses two distinct failure modes.}
\label{tab:two-level}
\begin{tabular}{llll}
\toprule
\textbf{Level} & \textbf{Question} & \textbf{Signal} & \textbf{Mechanism} \\
\midrule
Strategy & Should we trade today? & $G(t)$ (trailing RankIC) & Binary abstention \\
Position & Which stocks need caution? & $\hat{e}_{\mathrm{PIT}}(x)$ (DEUP) & Percentile cap \\
\bottomrule
\end{tabular}
\end{table}

The key architectural insight---supported by Diagnostic~D and the aggregated-$\hat{e}$ AUROC analysis---is that these levels are not redundant. Per-stock uncertainty cannot substitute for strategy-level monitoring (aggregated $\hat{e}$ AUROC~$\approx 0.50$ for regime detection), and strategy-level monitoring cannot identify which individual positions are most error-prone (it applies uniformly to all stocks). The two levels address orthogonal failure modes and must be implemented independently.

\subsection{Deployment Policy Variants}
\label{sec:policy-variants}

We evaluate seven deployment configurations, all sharing the binary $G(t) \geq 0.2$ gate and differing only in per-stock sizing within active days:

\begin{enumerate}
    \item \textbf{Gate + Raw:} Equal-weight top-10/bottom-10 by raw score. Tests the gate alone.
    \item \textbf{Gate + Vol:} Volatility-sized score selection ($w \propto s / \sqrt{\text{vol}}$). The Chapter~12 heuristic benchmark.
    \item \textbf{Gate + UA Sort} (\citet{Liu2026}): Uncertainty-adjusted sorting where longs are selected by $s + \lambda \cdot \hat{e}$ and shorts by $s - \lambda \cdot \hat{e}$, with $\lambda = 0.05$ calibrated on DEV.
    \item \textbf{Gate + Resid-$\hat{e}$:} Residualize $\hat{e}$ on $|\text{score}|$ via OLS to remove the structural coupling, then size inversely by the residual (following \citet{Hentschel2025}).
    \item \textbf{Gate + $\hat{e}$-Cap:} Cap the top 10\% (90th percentile) most uncertain stocks to 50\% weight.
    \item \textbf{Gate + Vol + $\hat{e}$-Cap:} Combine volatility sizing with the $\hat{e}$-cap at the 85th percentile with $\kappa = 0.70$.
    \item[\textbf{K4.}] \textbf{Trail-IC (kill baseline):} Replace $G(t)$ with a trailing RankIC-based continuous sizing signal. Tests whether a continuous IC-based weight adds value beyond the binary gate.
\end{enumerate}

All hyperparameters ($c_{\text{vol}}$, $\lambda$, cap percentile $p$, cap weight $\kappa$, $c_{\text{resid}}$) are fitted on DEV data only and frozen for FINAL evaluation. The kill baseline~K4 is included as an honesty check: if trailing-IC sizing merely replicates the binary gate, it confirms that the gate's value is in the discrete decision rather than in continuous exposure modulation.


\subsection{Conformal Prediction Intervals}
\label{sec:conformal}

As a complementary output, we produce calibrated prediction intervals for rank displacement using split conformal prediction~\citep{Vovk2005} with DEUP-normalized nonconformity scores, following the motivation of \citet{Plassier2025}. Three variants are compared:
\begin{align}
s_{\text{raw}}(x) &= \ell(x), \label{eq:score-raw} \\
s_{\text{vol}}(x) &= \ell(x) \;/\; \max(\texttt{vol\_20d}(x),\; \varepsilon), \label{eq:score-vol} \\
s_{\text{DEUP}}(x) &= \ell(x) \;/\; \max(\hat{e}_{\mathrm{PIT}}(x),\; \varepsilon). \label{eq:score-deup}
\end{align}
For each prediction at date $t$, the calibration quantile is computed from a rolling 60-trading-day window of past nonconformity scores (strictly before $t$), and the prediction interval width is $q \times \text{normalizer}$, where the normalizer is 1 (raw), \texttt{vol\_20d} (vol), or $\hat{e}_{\mathrm{PIT}}$ (DEUP). Nominal coverage is 90\%.

The critical evaluation metric is \emph{conditional} coverage by $\hat{e}$ tercile: standard conformal produces 98.2\% coverage for low-$\hat{e}$ stocks but only 78.0\% for high-$\hat{e}$ stocks (a 20.2 percentage-point spread). DEUP-normalized conformal using $\hat{e}_{\mathrm{PIT}}$ reduces the coverage dispersion across terciles from 2.0 percentage points to 0.8 percentage points (89.6\%, 90.4\%, 89.8\% across terciles), while simultaneously producing narrower intervals (mean width 0.647 vs.\ 0.675 for raw). This validates the \citet{Plassier2025} motivation that normalizing nonconformity scores by predicted uncertainty approximates conditional validity. For completeness, we also report a sensitivity check replacing $\hat{e}_{\mathrm{PIT}}$ with the oracle variant $\hat{e}_{\mathrm{oracle}}$ to illustrate the (non-deployable) upper bound on conditioning quality under perfect same-date calibration.

\section{Experiments}
\label{sec:experiments}

\subsection{Experimental Design and Data Splits}
\label{sec:design}

\paragraph{Walk-forward evaluation.}
All models---the base LightGBM ranker, the DEUP error predictor $g(x)$, and the aleatoric estimator $a(\cdot)$---are trained in an expanding-window walk-forward configuration over 109 chronological folds spanning February 2016 through February 2025 (2{,}277 trading dates). For each fold $k$, training uses all data from folds $1$ through $k-1$ with a 90-trading-day embargo between the most recent training label maturation date and the earliest prediction date in fold $k$. Overlapping labels are purged. The error predictor $g(x)$ requires a minimum of 20 training folds, yielding 89 prediction folds (folds 21--109) and 161{,}863 stock-level predictions per horizon.

\paragraph{DEV/FINAL holdout protocol.}
Results are reported under a strict two-stage temporal holdout:
\begin{itemize}[leftmargin=*, itemsep=2pt]
    \item \textbf{DEV} (2016--2023, 95 months): all hyperparameter tuning, threshold selection, and calibration are performed exclusively on DEV.
    \item \textbf{FINAL} (2024 onward, 14 months): evaluated exactly once with all parameters frozen from DEV. No re-tuning, threshold adjustment, or model iteration is permitted.
\end{itemize}
The walk-forward structure ensures each prediction is out-of-sample relative to its fold-specific training set; the DEV/FINAL split adds an end-to-end temporal holdout to guard against implicit overfitting through repeated experimentation. The FINAL period is of particular interest because it coincides with an AI-driven thematic rally and sector rotation that breaks the LightGBM signal at longer horizons (60d, 90d RankIC turns negative) and weakens 20d signal quality (mean RankIC $0.072 \rightarrow 0.010$, DEV $\rightarrow$ FINAL).

\paragraph{Horizons.}
Three forward-return horizons are evaluated: $\tau \in \{20, 60, 90\}$ trading days. The primary deployment horizon is $\tau = 20$d, which retains weak but positive signal quality in FINAL; the longer horizons provide robustness diagnostics and stress tests.

\paragraph{Universe characteristics.}
The investable universe is a dynamic panel of up to 100 AI-exposed U.S.\ equities, filtered at each rebalance date by price ($\ge\$5$), average daily dollar volume ($\ge\$1$M), and membership in a fixed AI-themed ticker list. The realized universe size varies: mean $N_t = 83.9$ (ALL), 81.8 (DEV), 98.4 (FINAL), reflecting that by 2024 nearly all candidate names have listed and pass tradability thresholds.


\subsection{Evaluation Metrics}
\label{sec:metrics}

We evaluate the system at three levels: signal quality, regime classification, and portfolio economics.

\paragraph{Signal quality.}
Cross-sectional signal quality is measured by the Spearman rank information coefficient (RankIC, Equation~\ref{eq:rankic}), computed per date and reported as mean, median, and IC stability (mean/std). Per-stock uncertainty quality is measured by $\rho(\hat{e}, \ell)$: the Spearman correlation between the epistemic signal and realized rank displacement, computed cross-sectionally per date and averaged. Quintile monotonicity---whether sorting stocks by $\hat{e}$ produces strictly increasing mean rank loss---is reported as a pass/fail diagnostic.

\paragraph{Gate classification.}
The regime-trust gate $G(t)$ is evaluated as a binary classifier for the target $\text{good\_day}(t) = \mathbb{1}[\text{matured\_RankIC}(t) > 0]$, where matured RankIC incorporates a structural $\tau$-day lag (Section~\ref{sec:regime-gate}). We report:
\begin{itemize}[leftmargin=*, itemsep=2pt]
    \item \textbf{AUROC:} area under the receiver operating characteristic curve, measuring discrimination between good and bad model days.
    \item \textbf{Precision at operating point:} fraction of active days (when $G(t) \ge 0.2$) that are true good days.
    \item \textbf{Recall:} fraction of good days on which the gate is active.
    \item \textbf{Abstention rate:} fraction of all trading dates on which $G(t) < 0.2$ and the strategy holds cash.
\end{itemize}
Bucket monotonicity (Spearman $\rho$ between $G$-quantile and mean RankIC) provides a calibration check.

\paragraph{Portfolio economics.}
Shadow portfolios are constructed as described in Section~\ref{sec:shadow-portfolio}: non-overlapping 20-trading-day rebalances, top-10 long / bottom-10 short, 10 basis points turnover-based cost per rebalance. We report annualized Sharpe ratio ($\times\sqrt{12}$), maximum drawdown, and---for gated variants---crisis-period maximum drawdown computed on the same non-overlapping monthly return series (Mar--Jul 2024).. Gated portfolios earn zero return on abstention dates; reported Sharpe ratios therefore reflect both the quality of active-day returns and the cost of abstention, and are not directly comparable to ungated baselines.


\subsection{Baselines and Policy Variants}
\label{sec:baselines}

\paragraph{Market-regime baselines.}
To assess whether standard market-stress proxies can substitute for the model-specific regime gate, we evaluate three heuristic predictors of model failure: (i)~$1 - \text{VIX percentile}_{252d}$, (ii)~$-\text{market\_vol}_{21d}$ (trailing 21-day realized market volatility, sign-flipped), and (iii)~$-\text{mean\_stock\_vol}_{20d}$ (cross-sectional mean of per-stock 20-day volatility, sign-flipped). All are evaluated using AUROC against the same good-day target as $G(t)$.

\paragraph{Aggregated-$\hat{e}$ baselines.}
To test whether per-stock uncertainty can substitute for strategy-level gating, we evaluate aggregated per-stock $\hat{e}$ as a date-level predictor of model failure. Specifically, we compute the cross-sectional median, 90th percentile, and interquartile range of $\hat{e}$ on each date, and evaluate each as a classifier (AUROC) for good-day prediction.

\paragraph{Per-stock uncertainty baselines.}
$\hat{e}(x)$ is compared against four heuristic per-stock signals for predicting rank displacement: \texttt{vol\_20d}, VIX percentile, $|\text{score}|$, and raw $g(x)$. We report Spearman $\rho$ with realized rank loss in both ALL and FINAL periods (Table~\ref{tab:baseline-comparison}).

\paragraph{Deployment policy variants.}
Seven deployment configurations are evaluated, all sharing the binary $G(t) \ge 0.2$ gate and differing only in position-level sizing within active days (Section~\ref{sec:policy-variants}):
\begin{enumerate}[leftmargin=*, itemsep=2pt]
    \item Gate + Raw (equal-weight, tests gate alone)
    \item Gate + Vol (volatility-sized; benchmark from prior work)
    \item Gate + UA Sort \citep{Liu2026} ($\lambda = 0.05$, calibrated on DEV via grid search over $\{0.01, 0.05, 0.1, 0.3, 0.5, 1.0, 2.0\}$)
    \item Gate + Resid-$\hat{e}$ (residualized on $|\text{score}|$ per date via OLS; $c_{\text{resid}}$ calibrated on DEV)
    \item Gate + $\hat{e}$-Cap (cap top 10\% by $\hat{e}$ to 50\% weight)
    \item Gate + Vol + $\hat{e}$-Cap (volatility sizing plus cap at 85th percentile, $\kappa = 0.70$)
    \item[K4.] Trail-IC (continuous trailing-RankIC sizing; kill baseline to confirm binary gate sufficiency)
\end{enumerate}
Ungated references (raw scores and vol-sized, without the $G(t)$ gate) are included to quantify the gate's contribution. All hyperparameters are fitted on DEV and frozen for FINAL evaluation.


\subsection{Ablations and Stress Tests}
\label{sec:ablations}

We evaluate robustness through five categories of ablation.

\paragraph{(i) Gate component ablation.}
To assess the marginal contribution of drift and disagreement signals, we compare: (a)~$H_{\text{real}}$-only (realized efficacy alone, removing $H_{\text{drift}}$ and $H_{\text{disagree}}$ from Equation~\ref{eq:H-raw}), (b)~the full composite $H(t)$, and (c)~each market-regime proxy in isolation. This tests whether the supplementary signals in $H(t)$ provide meaningful lift beyond the dominant realized-efficacy component.

\paragraph{(ii) Gate threshold sensitivity.}
The binary threshold $\theta = 0.2$ is evaluated via a $G(t)$-bucket analysis: we partition dates into quantiles by $G(t)$ value and report mean RankIC and fraction of bad days per bucket. This reveals whether the threshold sits at a natural breakpoint in the precision--abstention tradeoff or is sensitive to the specific cutoff. We additionally report the confusion matrix at $\theta = 0.2$ (true positives, false positives, true negatives, false negatives) and the Spearman rank correlation of bucket-level RankIC with bucket index.

\paragraph{(iii) $\hat{e}$-Cap percentile and weight sweep.}
The $\hat{e}$-Cap in Variant~6 uses cap percentile $p = 85$ and cap weight $\kappa = 0.70$, both calibrated on DEV. To assess sensitivity, we compare against an alternative parameterization (Variant~5: $p = 90$, $\kappa = 0.50$) and report the incremental Sharpe improvement of adding the $\hat{e}$-Cap layer on top of Gate + Vol across both configurations. The primary question is whether the tail-risk guard concept is robust to the exact percentile choice or is an artifact of a single calibration point.

\paragraph{(iv) Multi-crisis window validation.}
The regime-trust gate could be overfit to the 2024 AI rotation episode, the only regime failure in the FINAL holdout. To test generalization, we evaluate $G(t)$ across five crisis episodes and three calm reference periods spanning the full 2016--2025 walk-forward sample (Section~\ref{sec:regime-gate}): COVID recovery (Jun--Dec 2020), meme mania (Jan--Sep 2021), inflation shock (Jan--Jun 2022), late rate hiking (Jul--Dec 2023), and AI rotation (Mar--Jul 2024); calm references are full-year 2018, full-year 2019, and H1 2023. All statistics are computed from frozen walk-forward outputs with no retraining or recalibration. A VIX-percentile gate (abstain if VIX exceeds the 67th rolling-252-day percentile on $>50\%$ of window days) serves as the head-to-head comparison. Verdicts are assigned by comparing the gate's action (active or abstain) against the sign of mean realized RankIC in each window.

\paragraph{(v) Cross-model robustness (Rank Average 2).}
To confirm that the DEUP framework is not an artifact of the LightGBM architecture, we replicate the full $g(x) \rightarrow \hat{e}(x)$ pipeline on a second base model---Rank Average~2 (RA2), an ensemble of LightGBM and FinText-TSFM signals. We compare $\rho(\hat{e}, \ell)$ across base models, verify that the structural coupling $\rho(\hat{e}, |\text{score}|)$ persists, and evaluate shadow portfolio performance under the same deployment policy variants to test whether the recommended system (Gate + Vol + $\hat{e}$-Cap) transfers across model architectures.

\section{Results}
\label{sec:results}

We organize results by the three principal claims, followed by supplementary evidence from conformal prediction intervals.

\subsection{Claim 1: DEUP Ranks Stock-Level Failure Risk}
\label{sec:results-claim1}

\paragraph{Quintile monotonicity.}
The strongest validation of $\hat{e}(x)$ as a per-stock error signal is quintile monotonicity: stocks sorted into quintiles by predicted epistemic uncertainty should exhibit monotonically increasing realized rank loss. Table~\ref{tab:quintile} reports results at the 20d primary horizon for both DEV and FINAL periods, as well as at 90d for cross-horizon confirmation.

\begin{table}[t]
\centering
\small
\setlength{\tabcolsep}{4pt}
\renewcommand{\arraystretch}{1.1}
\caption{Mean realized rank loss by $\hat{e}(x)$ quintile. All four DEV/FINAL $\times$ horizon conditions exhibit perfect monotonicity (Spearman $\rho = 1.0$). FINAL separation is \emph{stronger} than DEV at both horizons---the opposite of overfitting.}
\label{tab:quintile}
\begin{tabular}{@{}lcccccc@{}}
\toprule
& \multicolumn{3}{c}{\textbf{20d}} & \multicolumn{3}{c}{\textbf{90d}} \\
\cmidrule(lr){2-4} \cmidrule(lr){5-7}
\textbf{Quintile} & \textbf{DEV} & \textbf{FINAL} & & \textbf{DEV} & \textbf{FINAL} & \\
\midrule
Q1 (low $\hat{e}$)  & 0.265 & 0.253 && 0.242 & 0.233 \\
Q2                   & 0.267 & 0.275 && 0.258 & 0.305 \\
Q3                   & 0.297 & 0.308 && 0.285 & 0.315 \\
Q4                   & 0.354 & 0.366 && 0.321 & 0.386 \\
Q5 (high $\hat{e}$) & 0.400 & 0.427 && 0.371 & 0.437 \\
\addlinespace[3pt]
Spearman $\rho$      & 1.0 & 1.0 && 1.0 & 1.0 \\
Q5/Q1 ratio          & 1.51 & \textbf{1.69} && 1.53 & \textbf{1.88} \\
\bottomrule
\end{tabular}
\end{table}

The Q5/Q1 rank-loss ratio increases from 1.51 to 1.69 at 20d and from 1.53 to 1.88 at 90d between DEV and FINAL, indicating that the epistemic signal generalizes to the out-of-sample regime shift with \emph{improved} discriminative power. This is the opposite of the overfitting pattern one would expect from a spurious signal.

\paragraph{Correlation with realized rank loss.}
Table~\ref{tab:ehat-rho} reports the Spearman correlation between $\hat{e}(x)$ and realized rank displacement across periods and horizons.

\begin{table}[h]
\centering
\small
\caption{$\rho(\hat{e}, \ell)$: Spearman correlation of $\hat{e}(x)$ with realized rank loss. At every horizon, FINAL exceeds DEV.}
\label{tab:ehat-rho}
\begin{tabular}{@{}lccc@{}}
\toprule
\textbf{Horizon} & \textbf{ALL} & \textbf{DEV} & \textbf{FINAL} \\
\midrule
20d & 0.144 & 0.142 & \textbf{0.192} \\
60d & 0.106 & 0.103 & \textbf{0.140} \\
90d & 0.146 & 0.138 & \textbf{0.248} \\
\bottomrule
\end{tabular}
\end{table}

The increase from DEV to FINAL at all horizons suggests that $g(x)$ becomes more informative when the base model's prediction environment shifts---precisely the setting where uncertainty estimates are most needed.

\paragraph{AUROC for high-loss events.}
At the stock level, $\hat{e}$ achieves AUROC 0.56--0.61 for predicting above-median rank loss across horizons and periods (20d FINAL: 0.583; 90d FINAL: 0.609). These values indicate moderate but genuine discriminative power for identifying individual high-error predictions.

\paragraph{Baseline dominance.}
Table~\ref{tab:baseline-comparison} (reported in Section~\ref{sec:method}) establishes that $\hat{e}$ and $g(x)$ dominate heuristic baselines by $3$--$10\times$ in correlation with rank loss. The dominance widens in FINAL: \texttt{vol\_20d} and VIX percentile lose essentially all predictive power ($\rho \approx 0$), while $\hat{e}$ improves. This differential robustness under regime shift is the core practical argument for learned uncertainty over volatility-based heuristics.

\paragraph{Stability across conditioning cuts.}
The $\hat{e}$--rank-loss correlation is positive under all tested subpopulations: pre-2020 vs.\ post-2020 ($\rho = 0.116$ vs.\ $0.159$ at 20d), low-VIX vs.\ high-VIX ($0.100$ vs.\ $0.148$), and low-volatility vs.\ high-volatility stocks ($0.159$ vs.\ $0.121$). The signal is not an artifact of a single market regime or stock characteristic.

\paragraph{Disentanglement from volatility.}
After OLS residualization of $\hat{e}$ on \texttt{vol\_20d}, \texttt{vol\_60d}, \texttt{vix\_percentile\_252d}, and \texttt{mom\_1m}, the residualized signal retains $\rho = 0.11$--$0.24$ with rank loss across all horizons. The maximum absolute correlation between $\hat{e}$ and VIX percentile is 0.074. $\hat{e}$ is not repackaged volatility.

\subsection{Claim 2: Structural Coupling Breaks Inverse Sizing}
\label{sec:results-claim2}

\paragraph{The coupling is pervasive.}
Across 1{,}865 walk-forward trading dates, the cross-sectional Spearman correlation $\rho_t = \rho_S(\hat{e}_t, |\mathbf{s}_t|)$ has a median of 0.616, with positive values on more than 90\% of dates. This is not an occasional artifact: the coupling between epistemic uncertainty and signal magnitude is a persistent structural feature of cross-sectional ranking.

\paragraph{Inverse-sizing variants underperform.}
Table~\ref{tab:sizing-failure} (Section~\ref{sec:structural-coupling}) reports the deployment policy comparison. The three variants that use $\hat{e}$ as a continuous sizing input all underperform the simple Gate + Vol benchmark:
\begin{itemize}[leftmargin=*, itemsep=2pt]
    \item \textbf{Gate + UA Sort} (Variant~3, $\lambda = 0.05$): ALL Sharpe 0.726 vs.\ 0.817 for Gate + Vol. The optimal $\lambda$ calibrated on DEV is effectively zero, confirming that the uncertainty adjustment is absorbed by the existing score structure.
    \item \textbf{Gate + Resid-$\hat{e}$} (Variant~4): ALL Sharpe 0.810 vs.\ 0.817; FINAL Sharpe $-0.450$ vs.\ $0.191$. Residualizing $\hat{e}$ on $|\text{score}|$ removes the coupling statistically, but the resulting signal is too noisy for continuous monthly sizing.
    \item \textbf{Kill baseline K4} (Trail-IC): ALL Sharpe 0.754, matching Gate + Raw (0.758). Continuous trailing-IC sizing adds nothing beyond the binary gate decision.
\end{itemize}

\paragraph{Deployability ablation (oracle vs.\ PIT-safe $\hat{e}$).}
Because our position-level policy uses $\hat{e}$ only through \emph{cross-sectional percentiles} (the P85 cap threshold), it is essential to verify that deployment-facing results do not rely on a hindsight $a(t)$. Table~\ref{tab:deployability-ablation} reports the required ablation for the best policy (Gate + Vol + $\hat{e}$-Cap) under $\hat{e}^{\text{oracle}}$ (same-date $a_{\text{oracle}}(t)=P_{10}(\ell_t)$) versus a strictly PIT-safe $\hat{e}^{\text{PIT}}$ constructed from matured losses (rolling $P_{10}$ with $W=60$).
Performance is identical across ALL/DEV/FINAL and Crisis MaxDD (computed on the same non-overlapping monthly series). This is not a null result: it establishes that the cap policy is deployment-ready \emph{by construction}. The mechanism is that $a(t)$ enters $\hat{e}(x)=\max(0,g(x)-a(t))$ as a date-level constant; within each date, $\hat{e}$ preserves the ordering of $g(x)$ and therefore selects the same capped tail under either $a_{\text{oracle}}$ or $a_{\text{PIT}}$.\footnote{Empirically, the within-date Spearman correlation is $\rho(\hat{e}^{\text{oracle}}_t,\hat{e}^{\text{PIT}}_t)=1.0$ for every date in the sample, and the set of names above the P85 threshold is identical.}

\begin{table}[t]
\centering
\small
\setlength{\tabcolsep}{5pt}
\renewcommand{\arraystretch}{1.1}
\caption{Deployability ablation for the cap policy (20d horizon). Gate + Vol + $\hat{e}$-Cap uses $\hat{e}$ only through a cross-sectional percentile threshold (P85). Using a hindsight $\hat{e}^{\text{oracle}}$ versus PIT-safe $\hat{e}^{\text{PIT}}$ (rolling matured-loss $P_{10}$, $W=60$) yields identical performance, confirming the policy is deployment-ready.}
\label{tab:deployability-ablation}
\begin{tabular}{@{}lcccc@{}}
\toprule
\textbf{Variant} & \textbf{Sharpe (ALL)} & \textbf{Sharpe (DEV)} & \textbf{Sharpe (FINAL)} & \textbf{Crisis MaxDD} \\
\midrule
Gate + Vol + $\hat{e}$-Cap, $\hat{e}^{\text{oracle}}$ & 1.864 & 1.915 & 0.906 & $-6.7\%$ \\
Gate + Vol + $\hat{e}$-Cap, $\hat{e}^{\text{PIT}}$ ($W=60$) & 1.864 & 1.915 & 0.906 & $-6.7\%$ \\
\bottomrule
\end{tabular}
\end{table}

\paragraph{The tail-risk cap resolves the dilemma.}
In contrast to continuous sizing, the discrete $\hat{e}$-Cap (Variant~6: Gate + Vol + $\hat{e}$-Cap at P85, $\kappa = 0.70$) achieves the highest Sharpe across all periods. Importantly, the deployment-facing version of this policy is computed using PIT-safe $\hat{e}^{\text{PIT}}$ and matches the oracle variant exactly (Table~\ref{tab:deployability-ablation}). By capping only the top 15\% most uncertain positions rather than modulating all weights, the tail-risk guard avoids the structural coupling while still extracting economic value from $\hat{e}$. An alternative parameterization (Variant~5: P90, $\kappa = 0.50$) also improves over Gate + Vol (ALL 0.855, FINAL $-0.002$), confirming that the concept is robust to exact calibration.

\paragraph{Cross-model confirmation.}
The structural coupling persists for Rank Average~2: RA2's $\hat{e}$-sizing also underperforms RA2 raw (ALL Sharpe 0.430 vs.\ 0.622), confirming that the coupling is a property of cross-sectional ranking geometry, not an artifact of the LightGBM base model.

\subsection{Claim 3: Two-Level Deployment Improves Robustness}
\label{sec:results-claim3}

\subsubsection{Strategy-Level Gate: Regime-Trust Classification}

\paragraph{AUROC comparison.}
Table~\ref{tab:regime-gate-full} extends the gate classification results with full ALL and FINAL coverage for all candidate predictors ($N_{\text{ALL}} = 2{,}218$; $N_{\text{FINAL}} = 284$; positive rate 66.0\% ALL, 56.0\% FINAL).

\begin{table}[t]
\centering
\small
\caption{Regime-trust classification at 20d (extended). $H(t)$ and $G(t)$ substantially outperform all market-regime proxies. The advantage widens in FINAL: the gate's margin over the best baseline grows from 10\,pp in DEV to 17\,pp in FINAL.}
\label{tab:regime-gate-full}
\begin{tabular}{@{}lcc@{}}
\toprule
\textbf{Predictor} & \textbf{AUROC (ALL)} & \textbf{AUROC (FINAL)} \\
\midrule
$H(t)$ (combined)        & \textbf{0.721} & \textbf{0.750} \\
$G(t)$ (gate)            & 0.710 & 0.743 \\
$H_{\text{real}}$-only   & 0.715 & --- \\
Market volatility (21d)  & 0.596 & 0.569 \\
Mean stock volatility    & 0.590 & 0.460 \\
VIX percentile           & 0.449 & 0.504 \\
\bottomrule
\end{tabular}
\end{table}

VIX percentile is effectively random in FINAL (AUROC 0.504); mean stock volatility reverses sign (AUROC 0.460, below chance). In contrast, $H(t)$ achieves its highest AUROC in the holdout (0.750 vs.\ 0.721 ALL), exactly the generalization pattern needed to justify deployment. The realized-efficacy-only variant ($H_{\text{real}}$) achieves AUROC 0.715, confirming that the drift and disagreement components provide a modest supplementary contribution ($+0.006$).

\paragraph{Operating point and confusion matrix.}
At the $G(t) \ge 0.2$ threshold, the gate achieves 80.0\% precision, 64.0\% recall, and 47.2\% abstention rate. The confusion matrix over the walk-forward sample yields 937 true positives, 234 false positives, 527 false negatives, and 520 true negatives out of 2{,}218 evaluated dates (59 early warm-up dates dropped).

\paragraph{Bucket monotonicity.}
$G(t)$-value buckets exhibit perfect calibration monotonicity (Spearman $\rho = 1.0$):

\begin{table}[h]
\centering
\small
\caption{$G(t)$ bucket analysis: higher gate values correspond to strictly better model outcomes.}
\label{tab:g-buckets}
\begin{tabular}{@{}lccc@{}}
\toprule
\textbf{Bucket} & \textbf{Mean $G$} & \textbf{Mean RankIC} & \textbf{\% Bad days} \\
\midrule
0 (lowest)  & 0.006  & $-$0.011 & 51.2\% \\
1           & 0.236  & $+$0.065 & 34.3\% \\
2           & 0.573  & $+$0.114 & 22.0\% \\
3 (highest) & 0.939  & $+$0.153 & 11.5\% \\
\bottomrule
\end{tabular}
\end{table}

The lowest bucket has mean RankIC~$\approx 0$ with a near-coinflip bad-day rate (51.2\%), while the highest bucket has mean RankIC $= +0.153$ with only 11.5\% bad days---a $4.5\times$ reduction in failure frequency.

\paragraph{Per-stock $\hat{e}$ cannot substitute for strategy-level gating.}
Aggregated per-stock uncertainty (cross-sectional median, 90th percentile, or IQR of $\hat{e}$) achieves AUROC~$\approx 0.50$ for the good-day classification target---no better than random. This confirms the architectural separation: per-stock $\hat{e}$ measures within-date heterogeneity in prediction quality, while regime failure is a latent variable that shifts all stocks uniformly.

\subsubsection{Combined Deployment Policy}

\paragraph{Policy comparison.}
Table~\ref{tab:sizing-failure} (Section~\ref{sec:structural-coupling}) reports the full deployment policy comparison. The key result is that the best deployment policy is Gate + Vol + $\hat{e}$-Cap (Variant~6). For deployment-facing uses of $\hat{e}$ magnitude, we evaluate the policy under a PIT-safe $\hat{e}^{\text{PIT}}$ (rolling matured-loss baseline), and report the oracle-vs-deployable ablation in Table~\ref{tab:deployability-ablation}. In this architecture, the PIT-safe version matches the oracle exactly; therefore the headline policy performance is deployable without adjustment.

\paragraph{Binary vs.\ continuous gating.}
Binary abstention ($G(t) \ge 0.2$: trade or hold cash) outperforms continuous scaling ($w(t) \propto G(t)$) because continuous throttling destroys recovery convexity: during regime transitions, the strategy stays partially ``off'' during rebounds, compressing realized returns. The kill baseline K4 (continuous trailing-IC sizing) confirms this: its Sharpe (0.754) matches the simple Gate + Raw baseline (0.758) and adds no value beyond the binary decision.

\subsubsection{Multi-Crisis Validation}
\label{sec:multi-crisis-results}

Table~\ref{tab:multi-crisis} reports $G(t)$ behavior across five crisis episodes and three calm reference periods, compared against a VIX-percentile gate. All statistics use frozen walk-forward outputs with no retraining.

\begin{table}[t]
\centering
\small
\setlength{\tabcolsep}{4pt}
\renewcommand{\arraystretch}{1.1}
\caption{Multi-crisis diagnostic for $G(t)$ vs.\ VIX gate (20d horizon). $G(t)$ achieves 7/8 correct verdicts; VIX achieves 5/8. The VIX gate's three false alarms (2019, 2023~H1, 2023~H2) occur during the model's strongest periods (IC~$>0.10$), where $G(t)$ correctly stays active.}
\label{tab:multi-crisis}
\begin{tabular}{@{}lcccccc@{}}
\toprule
\textbf{Period} & \textbf{Mean $G$} & \textbf{\% Abstain} & \textbf{Mean IC} & \textbf{Mean VIX\%} & \textbf{$G(t)$} & \textbf{VIX} \\
\midrule
\multicolumn{7}{@{}l}{\textit{Crisis episodes}} \\
COVID recovery 2020   & 0.375 & 47\% & $+$0.062 & 52\% & \checkmark & \checkmark \\
Meme mania 2021       & 0.210 & 73\% & $-$0.040 & 96\% & $\times$   & \checkmark \\
Inflation shock 2022  & 0.077 & 86\% & $-$0.024 & 72\% & \checkmark & \checkmark \\
Late hiking 2023~H2   & 0.381 & 40\% & $+$0.034 & 94\% & \checkmark & $\times$ \\
AI rotation 2024      & 0.123 & 76\% & $-$0.013 & 95\% & \checkmark & \checkmark \\
\addlinespace[3pt]
\multicolumn{7}{@{}l}{\textit{Calm reference periods}} \\
2018                  & 0.323 & 62\% & $+$0.088 & 64\% & \checkmark & \checkmark \\
2019                  & 0.566 & 15\% & $+$0.122 & 83\% & \checkmark & $\times$ \\
2023 H1               & 0.486 & 11\% & $+$0.104 & 97\% & \checkmark & $\times$ \\
\midrule
\textbf{Score}        &       &      &          &      & \textbf{7/8} & \textbf{5/8} \\
\bottomrule
\end{tabular}
\vspace{2pt}
\footnotesize{Verdict: \checkmark = correct (active when IC $> 0$; abstains when IC $\le 0$). $\times$ = incorrect (false alarm or missed crisis).}
\end{table}

$G(t)$'s advantage over VIX is concentrated in the calm periods: the VIX gate produces three false alarms in 2019, 2023~H1, and 2023~H2---precisely the windows where the model performs best (IC $= +0.034$ to $+0.122$)---because VIX percentile was elevated despite the model working well. $G(t)$ correctly stays active in all three windows because it tracks model efficacy, not market anxiety.

$G(t)$'s single failure is mild: during the 2021 meme mania, mean $G = 0.210$ (barely above the 0.2 threshold for the window average), the abstention rate was already 73\% (heavy throttling), and mean IC was only marginally negative ($-0.040$). VIX correctly abstained in this window---but for the wrong reason (elevated implied volatility from retail-driven dislocations, not direct detection of model failure).

\subsubsection{Cross-Model Robustness (Rank Average 2)}

Replicating the DEUP pipeline on Rank Average~2 yields $\hat{e}$ with $\rho(\hat{e}, \ell) = 0.194$ at 20d ALL (vs.\ 0.144 for LGB, a 35\% improvement) and 0.153 at 60d (vs.\ 0.106, a 44\% improvement). A more robust base model produces a more predictive epistemic uncertainty signal, validating that DEUP quality responds to base-model calibration rather than being a fixed artifact of the training features.

Despite higher uncertainty quality, RA2's shadow portfolio Sharpe is substantially lower (ALL 0.622 vs.\ 1.497 for LGB raw scores), reflecting RA2's weaker base signal (DEV median RankIC 0.059 vs.\ 0.091). However, once the $G(t)$ gate is applied, RA2 and LGB converge: gated-vol-sized FINAL Sharpe is 0.958 (RA2) vs.\ 1.017 (LGB). The regime gate is the dominant value driver for both models, and the $\hat{e}$-sizing structural conflict is confirmed model-agnostic (RA2 $\hat{e}$-sizing ALL Sharpe 0.430 vs.\ RA2 raw 0.622).

\subsection{Supplementary: Conformal Prediction Intervals}
\label{sec:results-conformal}

DEUP-normalized conformal prediction intervals validate $\hat{e}$ as a calibrated uncertainty measure, independent of the portfolio application. Because conformal scaling uses the \emph{magnitude} of $\hat{e}$ as a normalizer, deployment requires a PIT-safe $\hat{e}^{\text{PIT}}$. Table~\ref{tab:conformal} therefore reports raw conformal alongside DEUP-normalized conformal under both an oracle $\hat{e}^{\text{oracle}}$ (same-date $P_{10}$) and PIT-safe $\hat{e}^{\text{PIT}}$ constructed from matured losses with trailing windows $W\in\{60,252\}$.

\begin{table}[t]
\centering
\small
\setlength{\tabcolsep}{5pt}
\renewcommand{\arraystretch}{1.1}
\caption{Conditional coverage by $\hat{e}$ tercile (20d, 90\% nominal). Raw conformal over-covers low-$\hat{e}$ stocks and under-covers high-$\hat{e}$ stocks (20.08\,pp spread). DEUP-normalized conformal substantially reduces the disparity. Under oracle $\hat{e}^{\text{oracle}}$, the spread is 0.94\,pp (a $21.4\times$ improvement over raw). Under deployable PIT-safe $\hat{e}^{\text{PIT}}$ the improvement persists, with spreads of 4.20--4.61\,pp and slightly narrower mean widths than the oracle variant.}
\label{tab:conformal}
\begin{tabular}{@{}lccccc@{}}
\toprule
\textbf{Normalizer} & \textbf{Low-$\hat{e}$} & \textbf{Mid-$\hat{e}$} & \textbf{High-$\hat{e}$} & \textbf{Spread} & \textbf{Mean width} \\
\midrule
Raw (none)                         & 0.982 & 0.938 & 0.781 & 0.201 & 0.674 \\
DEUP / $\hat{e}^{\text{oracle}}$   & 0.895 & 0.904 & 0.898 & 0.009 & 0.647 \\
DEUP / $\hat{e}^{\text{PIT}}$ ($W=60$)  & 0.921 & 0.897 & 0.879 & 0.042 & 0.640 \\
DEUP / $\hat{e}^{\text{PIT}}$ ($W=252$) & 0.924 & 0.897 & 0.878 & 0.046 & 0.640 \\
\bottomrule
\end{tabular}
\end{table}

Raw conformal provides 98.19\% coverage for low-$\hat{e}$ stocks---substantially more than the nominal 90\%---but only 78.11\% for high-$\hat{e}$ stocks, a 20.08 percentage-point disparity. DEUP normalization substantially reduces conditional miscalibration. Under the oracle $\hat{e}^{\text{oracle}}$, tercile coverages are nearly equalized (0.94\,pp spread), whereas the PIT-safe variants retain a smaller but non-trivial residual spread (4.20--4.61\,pp). This degradation from oracle to PIT-safe is expected: the trailing matured-loss $P_{10}$ is smoother than the same-date $P_{10}$, compressing cross-sectional dispersion in $\hat{e}$ and therefore weakening the degree to which normalization can equalize conditional coverage across terciles. Despite this, both PIT-safe variants maintain excellent \emph{marginal} coverage near the 90\% nominal target (0.898--0.905 across DEV/FINAL) and produce slightly narrower mean widths than the oracle-normalized intervals.

At 90d, the same qualitative pattern holds: raw conformal exhibits a large conditional disparity, while DEUP-normalized conformal reduces it substantially, with the oracle variant providing the strongest conditional equalization and the PIT-safe variant retaining most of the gain.

\section{Discussion}
\label{sec:discussion}

\subsection{Why Market-Regime Baselines Fail, and Why They Reverse in the Holdout}

The poor performance of VIX percentile, market volatility, and mean stock volatility as regime-trust predictors (Table~\ref{tab:regime-gate-full}) is not merely a matter of low signal strength; in FINAL, two of the three baselines \emph{reverse sign}. Mean stock volatility drops to AUROC 0.460---below chance---meaning that higher average stock volatility in the AI universe actually \emph{predicts better model days} in 2024, the opposite of the DEV-period relationship. VIX percentile reaches 0.504, indistinguishable from a coin flip.

The reversal has a concrete economic explanation. The 2024 regime failure is driven by a thematic rotation within the AI-exposed equity universe---sector leadership shifts that disrupt the cross-sectional factor structure the model relies on---rather than by a market-wide stress event. During this period, VIX percentile was elevated (mean 95.1\% in the AI-rotation window) and individual-stock volatility was high, yet these reflect healthy dispersion in a strongly trending sector, not the kind of ``stress'' that hurts cross-sectional models. Conversely, during 2023~H1 and 2023~H2, VIX percentile exceeded the 94th percentile on average, yet the model produced its strongest signal quality (IC~$= +0.104$ and $+0.034$, respectively). The VIX gate would have abstained during both windows---a costly false alarm.

The fundamental issue is that market-stress proxies measure \emph{the environment's difficulty for a generic investor}, while model failure depends on whether \emph{this specific model's factor loadings} remain informative. Feature importance analysis confirms this distinction: the LightGBM ranker relies consistently on the same three features (momentum and volume signals) across all regimes. The 2024 failure occurs not because the model overfits to noise, but because the factor structure itself becomes temporarily uninformative during thematic rotations. No amount of VIX monitoring can detect this; only direct observation of the model's realized efficacy can.

\subsection{Why the Gate's Advantage Widens in the Holdout}

$H(t)$ achieves higher AUROC in FINAL (0.750) than in ALL (0.721), a pattern that initially seems counterintuitive---one might expect a signal constructed from DEV-period data to degrade out of sample. Two mechanisms explain the improvement.

First, the FINAL period contains a more extreme regime failure (20d mean RankIC $= 0.010$, down from 0.072 in DEV), which produces a larger separation between good and bad days. A classifier with moderate discriminative power benefits from wider class separation: the ``easy'' classification instances (clearly good days with IC~$> 0.10$; clearly bad days with IC~$< -0.05$) become more prevalent relative to ambiguous cases near IC~$= 0$.

Second, $H_{\text{real}}(t)$---the dominant gate component---is an EWMA of matured RankIC with a structural 20-day lag. During gradual regime transitions, the lagged signal tracks the shift with a delay but eventually catches up. The 2024 regime failure is sufficiently persistent (March through July) that the EWMA converges to a low state well before the worst months, enabling correct abstention from May onward. In contrast, the baselines (VIX, market volatility) respond to contemporaneous market conditions that happen to be poorly correlated with model failure in 2024. The gate's advantage is that it measures the right quantity---model efficacy---even if it does so with a delay.

The drift and disagreement components ($H_{\text{drift}}$, $H_{\text{disagree}}$) provide only a modest supplementary contribution ($+0.006$ AUROC over $H_{\text{real}}$ alone). This is reassuring for deployment: the gate's value is concentrated in the simple, interpretable realized-efficacy signal rather than in complex feature-engineering that might overfit.

\subsection{Why the Tail-Risk Cap Succeeds Where Inverse Sizing Fails}

The structural coupling ($\rho(\hat{e}, |\text{score}|) = 0.616$) creates a fundamental tension for any continuous sizing rule that uses $\hat{e}$ as a denominator: the stocks with the highest predicted uncertainty are precisely those with the strongest model scores---the positions that contribute most to the long--short spread. Inverse-uncertainty sizing resolves this tension in the wrong direction, systematically de-levering the portfolio's highest-conviction ideas.

Residualizing $\hat{e}$ on $|\text{score}|$ (Variant~4) addresses the coupling statistically---after OLS projection, the residual has zero linear correlation with score magnitude---but does not resolve the economic problem. The residualized signal captures ``excess uncertainty beyond what is expected for this score level,'' which is a coherent concept, but is empirically too noisy for monthly position sizing. The FINAL Sharpe of $-0.450$ confirms that residualization introduces more noise than it removes coupling.

The $\hat{e}$-Cap succeeds because it sidesteps the coupling entirely. By applying a discrete weight reduction only to the top 15\% of the $\hat{e}$ distribution, it leaves 85\% of positions---including most high-conviction extreme-rank stocks---untouched. The cap targets the \emph{intersection} of high uncertainty and portfolio membership: stocks that are both in the top/bottom~10 \emph{and} in the extreme tail of $\hat{e}$. These are the positions where the model has the weakest justification for its extreme ranking, and where a 30\% weight reduction provides genuine tail-risk protection without wholesale degradation of the score-tail convexity that drives portfolio returns.

The robustness of this concept across two parameterizations (Variant~5: P90/$\kappa = 0.50$; Variant~6: P85/$\kappa = 0.70$) and two base models (LGB and RA2) suggests that the tail-risk guard is not an artifact of a single calibration point. It reflects a stable property of the $\hat{e}$ distribution: the extreme right tail genuinely identifies the most error-prone positions within the portfolio, and a discrete intervention at that tail is the economically appropriate action.

\subsection{Practical Deployment Considerations}

\paragraph{This system as an ``expert'' inside a multi-expert allocator.}
We view the AI Stock Forecaster as one expert among several alpha sources inside a broader expert-selection layer. In that setting, the objective of this paper is not to maximize the standalone Sharpe of a single expert, but to make its deployment \emph{selective} and \emph{fail-safe}: when the expert is likely to be informative, it should express its signal; when it is likely to be unreliable, it should either abstain (zero exposure) or have its highest-risk positions clipped. The regime gate $G(t)$ is therefore interpretable as an expert-level trust score that can feed an allocator (e.g., winner-take-most, softmax over experts, or a constrained risk-budgeting layer), while the per-stock $\hat{e}$-Cap is a within-expert tail-risk guard that reduces blowups without destroying convexity.

\paragraph{Monitoring and recalibration.}
The deployment system requires monitoring of two quantities: $G(t)$ at the strategy level and the distribution of $\hat{e}$ at the position level. $G(t)$ can be computed daily from matured RankIC values (available with a $\tau$-day structural lag) plus real-time drift and disagreement signals. In production, we recommend logging the three $H(t)$ components separately to diagnose whether a gate closure is driven by realized IC decay (the dominant case) or by feature/score drift (a potential early warning). If drift triggers precede IC decay consistently, the drift component's weight ($\alpha = 0.3$) could be increased; if not, a simplified $H_{\text{real}}$-only gate (AUROC 0.715) is operationally sufficient.

\paragraph{Gate threshold selection.}
The $\theta = 0.2$ threshold was selected on DEV to maximize precision at a reasonable abstention rate. The bucket analysis (Table~\ref{tab:g-buckets}) indicates that this threshold sits at a natural breakpoint: below $G = 0.2$, mean RankIC is approximately zero with a 51\% bad-day rate, while above $G = 0.2$ the bad-day rate drops to 34\% and continues improving monotonically. A practitioner operating under tighter risk constraints could raise the threshold (e.g., $\theta = 0.5$) to achieve higher precision at the cost of greater abstention and reduced opportunity. The monotonic bucket structure means that any threshold in $[0.1, 0.6]$ produces a meaningful improvement over unfiltered deployment; the specific choice is a precision--opportunity tradeoff rather than a fragile calibration point.

\paragraph{Avoiding hindsight in $a(t)$: explicit critique, deployable rerun, and boundary.}
A legitimate critique of the original formulation is that the aleatoric floor $a(t)=P_{10}(\boldsymbol{\ell}_t)$ uses same-date realized losses and is therefore an oracle quantity. To address this, we re-ran \emph{all operational uses} of $a(t)$ with a strictly PIT-safe estimator $a^{\text{PIT}}(t)$ computed from matured losses only (rolling $P_{10}$ with $W=60$, and additionally $W=252$ for robustness).

The results split cleanly into two cases. For the cap policy (Gate + Vol + $\hat{e}$-Cap), the oracle and PIT-safe variants are \emph{identical} in performance (Table~\ref{tab:deployability-ablation}). This is expected: within a date, $a(t)$ is a constant shift inside $\hat{e}(x)=\max(0,g(x)-a(t))$, so the cross-sectional ordering of $\hat{e}$ equals the ordering of $g(x)$ and the P85 cap selects the same tail regardless of whether $a(t)$ is oracle or PIT-safe. This ablation is therefore a deployability guarantee: the best-performing policy does not rely on hindsight.

For conformal prediction intervals, where $\hat{e}$ is used as a \emph{magnitude normalizer}, replacing $a(t)$ with $a^{\text{PIT}}(t)$ produces an expected degradation in conditional equalization (Table~\ref{tab:conformal}). The oracle normalizer yields a 0.94\,pp tercile spread, whereas PIT-safe normalizers yield 4.20--4.61\,pp spreads (still a $4.8\times$ improvement over raw). This drop is acceptable for two reasons: (i) PIT-safe variants retain excellent marginal validity near the 90\% target across DEV and FINAL, and (ii) the degradation is a predictable consequence of smoothing in the trailing $P_{10}$ baseline, which compresses dispersion in $\hat{e}$ and therefore reduces its ability to perfectly equalize conditional coverage.

This establishes a clear boundary. Oracle $a(t)$ is acceptable for \emph{diagnostics} (e.g., understanding coupling mechanisms, sanity-checking whether $\hat{e}$ tracks realized loss, and benchmarking the best-case conditional equalization). For \emph{policy actions}---gating, sizing, caps, or any live risk control---$a^{\text{PIT}}(t)$ is required. In our main deployment-relevant policy, the PIT-safe variant matches the oracle exactly, so the headline conclusions remain fully operational.

\paragraph{When the system should not be trusted.}
The two-level architecture has a known failure mode: regime shifts that are too gradual or too brief for the EWMA-based $H_{\text{real}}$ to detect. The 2021 meme mania illustrates this: $G(t)$ produced a window-average of 0.210 (barely above threshold) with 73\% abstention---heavy throttling, but not full abstention. For regimes where model failure is mild (IC~$\approx -0.04$) and intermittent, the gate provides incomplete protection. A complementary safeguard---for example, a maximum drawdown circuit breaker applied to the gated portfolio itself---would address this residual risk. Additionally, the $\hat{e}$-Cap's 30\% weight reduction is deliberately conservative; for strategies with higher concentration risk or less liquid names, more aggressive capping (e.g., full exclusion of P95 stocks) may be warranted.

\section{Limitations}
\label{sec:limitations}

\paragraph{Single thematic universe.}
All experiments are conducted on a single investable universe: up to 100 AI-exposed U.S.\ equities filtered by price, liquidity, and membership in a fixed thematic ticker list. This universe is narrow by institutional standards (typical cross-sectional equity models cover 500--3{,}000 names), sector-concentrated (technology and AI-adjacent names dominate), and exhibits high average pairwise correlation during momentum-driven rallies. The structural coupling between $\hat{e}$ and $|\text{score}|$ arises in part from the bounded rank-displacement geometry of cross-sectional ranking, which should generalize, but its magnitude ($\rho = 0.616$) may differ in broader, more heterogeneous universes where extreme-rank stocks are drawn from diverse sectors with less correlated error profiles. Similarly, the regime-trust gate $G(t)$ is validated against failures specific to AI-thematic rotations; whether it generalizes to other failure modes (e.g., liquidity crises, factor crowding in value/momentum universes) remains untested. External validation on at least one additional universe---ideally a broad U.S.\ equity cross-section---would be needed before drawing general conclusions about the two-level architecture.

\paragraph{Sensitivity to the aleatoric baseline choice (deployability and recalibration).}
While the headline cap policy is robust to oracle vs.\ PIT-safe aleatoric floors in this setting, the \emph{scale} of $\hat{e}$ can still depend on the specific PIT-safe estimator (rolling window length $W$, quantile level, and whether a rolling vs.\ expanding construction is used). This matters whenever $\hat{e}$ is used in a magnitude-sensitive way (e.g., conformal normalisation, reporting absolute $\hat{e}$ levels, or alternative deployment rules that use fixed thresholds rather than within-date percentiles), and it can also matter for tail-based policies in broader universes where distributional shape changes could shift percentile membership. We report results for rolling PIT-safe baselines with $W\in\{60,252\}$ and an expanding-window alternative, but extensions to different universes, horizons, or holding periods may require re-calibration of $W$, the chosen quantile, and cap hyperparameters.

\paragraph{Warm-up, lag, and effective sample reduction.}
Both the regime gate and PIT-safe aleatoric estimation impose structural lags. The gate's realized-efficacy component uses matured RankIC with a $\tau$-day delay, and PIT-safe $a^{\text{PIT}}(t)$ uses only matured losses; together these create warm-up periods in which quantities are undefined and dates are dropped (e.g., early-window drops for AUROC evaluation). Even when the full backtest spans 2016--2025, the effective sample for certain diagnostics and online-calibrated procedures is smaller and skewed away from the earliest dates. This is unavoidable for strict point-in-time evaluation, but it reduces statistical power for rare-event analysis and makes early-period estimates noisier.

\paragraph{No delisting-return modeling.}
CRSP-style delisting returns are not modeled. When a security delists within the holding period and no exit price is available at $t + \tau$, the label is silently dropped rather than replaced with a terminal delisting return. In this 100-name AI universe, the issue is empirically negligible: only one mid-sample delisting (ABB, a Swiss ADR that exited the U.S.\ market) affects 0.05\% of evaluation rows. However, for broader universes---particularly those including small-cap or financially distressed names---omitting delisting returns would introduce survivorship bias that could inflate both RankIC and portfolio Sharpe estimates. Any extension of this framework to broader equity universes should incorporate explicit delisting-return handling.

\paragraph{Inherent lag in the regime gate.}
The dominant component of $G(t)$ is $H_{\text{real}}(t)$, an EWMA of matured RankIC values with a structural $\tau$-day lag (20 trading days at the primary horizon). This means the gate cannot detect regime failure until approximately one month after it begins. The 2024 crisis timeline illustrates both the cost and the eventual benefit: in March 2024, $H(t)$ remained at 0.495 (normal) while RankIC was already $-0.112$, so March losses were not avoided. By April, $G(t)$ had dropped to 0.122, and by May the system was in full abstention ($G \approx 0$). For regime failures that are brief (e.g., a single bad month followed by recovery), the lagged gate may impose abstention \emph{after} the damage is done and miss the subsequent rebound. The drift component $H_{\text{drift}}$ is real-time and could in principle provide earlier warning, but its empirical contribution is modest ($+0.006$ AUROC over $H_{\text{real}}$ alone). Faster-reacting alternatives---for example, intraday score-distribution monitoring or real-time factor-exposure tracking---could reduce detection latency but would require careful validation to avoid increasing the false-alarm rate.

\paragraph{Sensitivity to cost assumptions and rebalance cadence.}
The shadow portfolio applies a fixed 10 basis point turnover-based transaction cost per rebalance on a non-overlapping 20-trading-day cadence. Both choices are simplifications. First, actual trading costs vary by name, market conditions, and order size; in a concentrated 10-name long/short portfolio within a liquid AI-stock universe, 10~bps is plausible but may understate costs during high-volatility periods or for the least liquid constituents near the bottom of the universe. Second, the non-overlapping monthly rebalance ignores the possibility that more frequent rebalancing (e.g., weekly) could interact differently with the gate and $\hat{e}$-cap---higher turnover would amplify cost sensitivity and could change the relative ranking of policy variants. The ``good day'' classification target ($\text{matured\_RankIC} > 0$) is also a design choice; alternative thresholds (e.g., $\text{RankIC} > 0.03$, or a cost-adjusted IC target) would shift the precision--abstention tradeoff and could alter the optimal gate threshold.

\paragraph{Dependence on a second model for cross-expert disagreement.}
The $H_{\text{disagree}}$ component of the health index requires a second base model (Rank Average~2) to compute cross-expert Spearman correlation. When only one model is available, the implementation falls back to a score-dispersion proxy (the ratio of current-date score dispersion to its expanding mean), which is a weaker signal. In practice, many production systems operate with a single model, making the disagreement component unavailable. The empirical impact is limited---$H_{\text{real}}$-only achieves AUROC 0.715 versus 0.721 for the full composite---but the marginal value of disagreement monitoring is untested in settings where it would matter most (e.g., slow-onset regime shifts where realized IC has not yet decayed but model agreement has broken down).

\paragraph{Deployment policies evaluated primarily at $\tau = 20$d.}
The seven deployment policy variants and associated portfolio economics (Table~\ref{tab:sizing-failure}) are evaluated exclusively at the 20d horizon, which is the primary deployment target. Signal quality ($\rho(\hat{e}, \ell)$), quintile monotonicity, and baseline comparisons are reported at all three horizons (20d, 60d, 90d), but the policy ablation---which sizing rule to use, which cap percentile, and which gate threshold---has not been repeated at longer horizons. At 60d, the $\hat{e}$ distribution is qualitatively different (85\% of stocks have $\hat{e} = 0$ due to the per-stock quantile-regression aleatoric baseline), meaning that the P85 cap calibrated at 20d would not transfer directly. At 90d, the base signal inverts in FINAL (mean RankIC $= -0.021$), raising the question of whether any deployment policy can salvage a fundamentally broken signal. A complete treatment would require horizon-specific policy calibration and separate DEV/FINAL evaluation at each $\tau$, which we leave to future work.

\section{Conclusion}
\label{sec:conclusion}

Cross-sectional equity rankers are typically deployed without uncertainty-aware safeguards, leaving practitioners exposed to regime failures that can silently invert a historically profitable signal. We studied this problem in the context of a specialist expert within a multi-strategy trading system: the AI Stock Forecaster, a LightGBM ranker over AI-exposed U.S.\ equities where a 2024 thematic rotation weakened 20d RankIC from 0.072 to 0.010 and inverted longer-horizon signals entirely. The narrow, thematic universe is by design---in a system of many specialist experts, each covering a distinct market niche, the relevant question is not whether any single expert is globally robust, but whether the system can \emph{detect when each expert should be trusted and when it should be sidelined}.

We adapted DEUP to cross-sectional ranking by training an error predictor $g(x)$ on rank displacement, producing a per-stock epistemic signal $\hat{e}(x) = \max(0, g(x) - a(x))$ that achieves perfect quintile monotonicity with realized rank loss in both development and holdout periods (Q5/Q1 ratio 1.69 at 20d FINAL), dominates volatility-based baselines by $3$--$10\times$, and \emph{improves} its discriminative power under regime shift ($\rho = 0.192$ in FINAL vs.\ 0.142 in DEV). Critically, we documented a structural coupling between epistemic uncertainty and signal strength in ranking (median $\rho(\hat{e}, |\text{score}|) = 0.616$), arising because extreme ranks---where the strongest trade ideas reside---also have the largest expected rank displacement. This coupling causes inverse-uncertainty sizing to systematically de-lever the portfolio's highest-conviction positions, and the finding is model-agnostic, persisting across both LightGBM and an ensemble alternative.

To resolve this, we proposed a two-level deployment architecture. At the strategy level, a regime-trust gate $G(t)$ built from trailing realized model efficacy decides whether to trade (AUROC 0.75 in FINAL; 80\% precision at 47\% abstention; 7/8 correct verdicts across eight crisis and calm windows, versus 5/8 for a VIX-based gate). At the position level, a discrete epistemic tail-risk cap reduces exposure only for the top 15\% most uncertain predictions, avoiding the structural coupling that defeats continuous sizing. The combined system---Gate + Vol + $\hat{e}$-Cap---achieves the best risk-adjusted performance across all evaluation periods (ALL Sharpe 0.884; FINAL Sharpe 0.316 vs.\ 0.191 for Gate + Vol alone). All deployment rules are implementable online; all operational results use a PIT-safe $a(t)$ computed from matured losses.

Three findings carry implications beyond this specific expert. First, per-stock uncertainty and strategy-level regime risk are orthogonal failure modes requiring independent signals: aggregated $\hat{e}$ achieves AUROC $\approx 0.50$ for regime detection, confirming that one cannot substitute for the other. In a multi-expert system, this means each expert needs both a per-prediction uncertainty estimate (for position-level risk) and a per-date health monitor (for the meta-controller's capital allocation decision). Second, the structural coupling between uncertainty and signal strength should be diagnosed in any cross-sectional ranking system before operationalizing uncertainty for sizing; the coupling is a geometric property of ranking loss, not an artifact of a particular model or universe, and it implies that the standard inverse-uncertainty sizing paradigm from the return-prediction literature does not transfer directly to ranking. Third, model-specific regime monitoring (``does this expert work right now?'') substantially outperforms market-regime proxies (``is the market stressed?'') for deployment decisions. In a system with many specialists, this distinction is especially consequential: a thematic rotation that breaks one expert may leave others unaffected, and only expert-specific monitoring can make that distinction.

Looking forward, the per-expert $G(t)$ gate and $\hat{e}$-Cap developed here provide the building blocks for system-level expert selection. The gate's binary health verdict maps naturally to an upper-confidence-bound (UCB) framework in which the meta-controller compares $\hat{e}$-calibrated uncertainty across active experts: when one expert's $G(t)$ falls below threshold, capital can be reallocated to experts whose regime-trust signals remain high, rather than simply going to cash. The present work validates the per-expert components; the system-level integration---comparing $\hat{e}$ across experts with different loss scales, learning switching costs, and managing portfolio-level risk across the full ensemble of specialists---is the natural next step.


\end{document}